%% file: arxiv.tex
\pdfoutput=1
\documentclass[10pt,twocolumn,letterpaper]{article}

\usepackage[pagenumbers]{cvpr} %

\usepackage{cuted} %

\usepackage{currfile} %

\usepackage{caption} %

\usepackage{microtype}

\usepackage{xcolor}
\usepackage{colortbl} %
\newcommand{\oursrow}{\rowcolor{gray!12}}
\newcommand{\method}{CATCH}          %
\newcommand{\projecturl}{https://dinnno.github.io/catch}  %
\newcommand{\samthree}{SAM~3.1}    %
\input{numbers}                                %

\definecolor{cvprblue}{rgb}{0.21,0.49,0.74}
\usepackage[pagebackref,breaklinks,colorlinks,allcolors=cvprblue]{hyperref}

\title{CATCH: Contact-Aware Trajectory Diffusion for Coupled Hand-Object Reconstruction from Egocentric Video}  %

\author{
Sangjun Noh\textsuperscript{1} \quad Jongwon Kim\textsuperscript{2} \quad Dongwoo Nam\textsuperscript{2} \quad Donghyun Hwang\textsuperscript{1,\textdagger} \quad Kyoobin Lee\textsuperscript{2,\textdagger}\\[0.3em]
{\normalsize \textsuperscript{1}Korea Institute of Science and Technology (KIST)}\\
{\normalsize \textsuperscript{2}Gwangju Institute of Science and Technology (GIST)}
\\[0.3em]
{\tt\small \{sangjun7, donghyun\}@kist.re.kr \quad kyoobinlee@gist.ac.kr}
}

\begin{document}
\maketitle
{\renewcommand{\thefootnote}{\textdagger}\footnotetext{Co-corresponding authors.}}
\begin{strip}
\centering
\includegraphics[width=\linewidth]{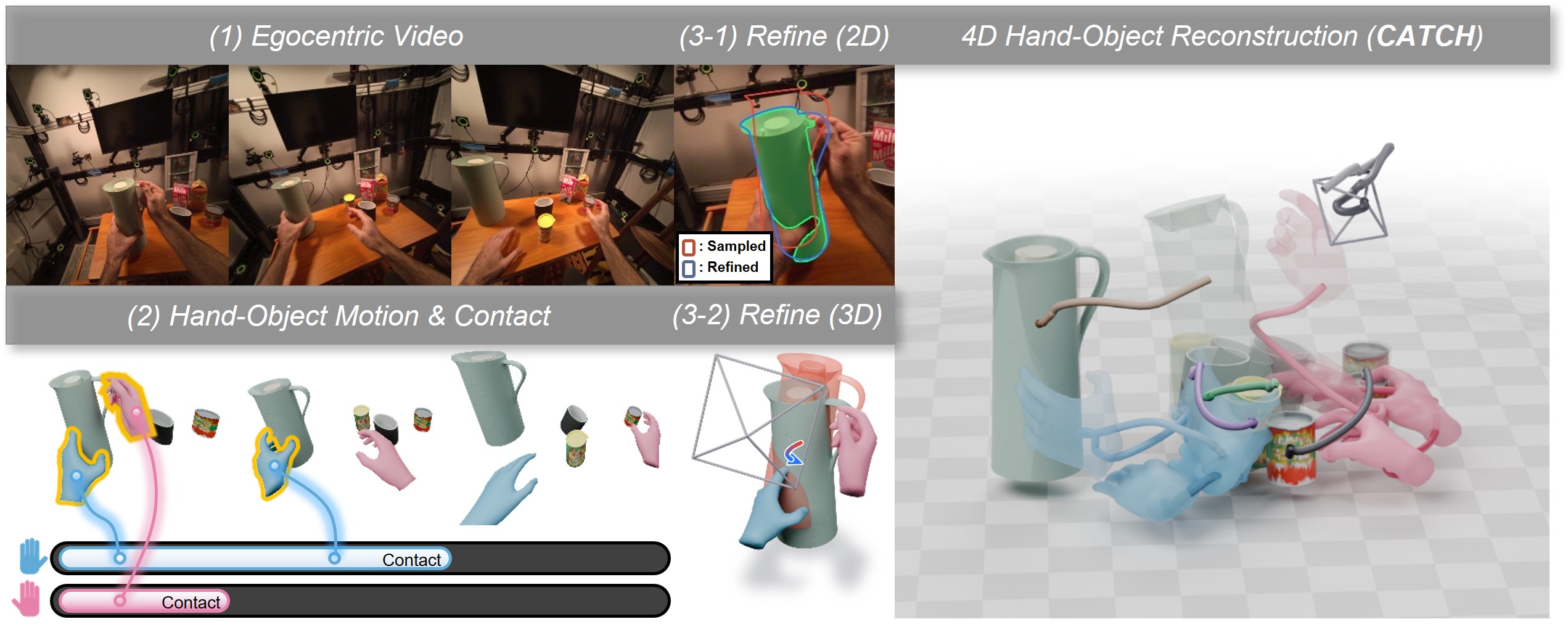}
\captionsetup{hypcap=false}
\captionof{figure}{\textbf{\method{} reconstructs 4D hand--object interaction from egocentric video (1).} From frozen foundation features of the video, a diffusion prior refines the hand track and samples the 6-DoF object trajectory and per-hand contact (2), with the contact of each hand shown as a bar. A short refinement then fits the sampled object to the observed mask (3), shown on the input frame (3-1; red: sampled, blue: refined, green: mask) and in 3D (3-2; red: sampled). Right: the resulting 4D reconstruction.}
\captionsetup{hypcap=true}
\label{fig:teaser}
\end{strip}

\begin{abstract}
Reconstructing how two hands and a manipulated object move together in 3D is central to understanding hand--object interaction (HOI) and to transferring human manipulation to robots.
For robots learning from human videos, the object trajectory is particularly important.
While hand estimators already perform well, the object is often occluded by the hands and remains hard to track.
Single-frame methods lose consistency when chained over a video.
Generative trajectory priors keep the interaction plausible over time, but rely on separate object pose estimators or external contact labels to locate the object.
These estimators ignore how hands and objects move together and often fail under such occlusion.
We propose \textbf{\method{}}, a diffusion prior that locates the object in egocentric video without an object pose estimator, given the object template, camera trajectory and visible masks.
Conditioned on frozen 3D geometry and appearance features from visual foundation models, \method{} refines an off-the-shelf hand track while jointly generating the 6-DoF object trajectory and per-hand contact.
A short refinement guided by the sampled contact then aligns this initial estimate with the masks, without an external contact model.
\method{} achieves state-of-the-art object pose accuracy on HOT3D while improving its input hand track, and transfers to H2O without fine-tuning.
We further demonstrate that its object trajectories more than double the success rate of physics-based robot retargeting over a dedicated object pose estimator.
\iftoggle{cvprfinal}{Project page: \url{\projecturl}.}{}
\end{abstract}

\section{Introduction}
\label{sec:intro}
Reconstructing hand--object interaction (HOI) in 4D, the 3D motion of the hands and the object over time, is a basis for understanding human activity and for teaching robots from human demonstrations.
Such an interaction is defined not only by how the hands move but also by what they do to the object~\cite{ehsani2020force}.
When robots learn from human videos, the object trajectory is particularly important, because it shows what the demonstration achieved rather than only how the hands moved~\cite{park2025learning,ponimatkin2025object,chen2026manipulation}.
Hand pose and motion estimation has been studied extensively and is already accurate~\cite{pavlakos2024hamer,potamias2025wilor,zhang2025hawor,yu2025dynhamr}, so we build on it and focus on how the object moves with the hands.
We therefore aim to reconstruct from egocentric video the world-space motion of both hands, the 6-DoF trajectory of the object and when each hand is in contact with it.

This task is challenging because the hands and the object occlude each other throughout the interaction, and the object usually moves only while it is held, so estimating hands~\cite{zhang2025hawor,potamias2025wilor} and objects~\cite{wen2024foundationpose} separately ignores their coupling.
Because the object moves with the hand, the hand track is a reasonable reference for locating the object.
Single-frame methods estimate the hand and the object jointly~\cite{hasson2019obman,cao2021mow,ye2022ihoi}, but chained over a video, their estimates do not keep the interaction consistent.
Recent methods instead learn generative priors over hand--object trajectories~\cite{ye2026whole,guo2026dynamichoi,fu2026egograsp}, given a known object template~\cite{ye2026whole,guo2026dynamichoi} and, as is common on headsets with SLAM, a known camera trajectory~\cite{ye2026whole}.
Such priors keep the interaction plausible over time, but they do not locate the object in the video by themselves.
WHOLE~\cite{ye2026whole} conditions its prior on the template and a hand track, and brings in the object masks and the contact labels it queries from a vision--language model only through guidance at every sampling step.
DynamicHOI~\cite{guo2026dynamichoi} and EgoGrasp~\cite{fu2026egograsp} instead start from a separate object pose estimate and refine it.
Either way, the object pose depends on external models that ignore how hands and objects move together and often fail under heavy hand occlusion, which also limits robot learning from human videos.

To overcome this, we propose \method{}, a video-conditioned diffusion prior that refines an off-the-shelf hand track while jointly sampling the 6-DoF object trajectory and per-hand contact from the RGB video, given the same template and camera and the visible object masks (\cref{fig:teaser}).
Our key insight is that frozen visual foundation models already provide the spatial cues that previous pipelines delegate to external estimators: $\pi^3$~\cite{wang2025pi3} features describe the 3D layout of the scene under a moving camera, and DINOv2~\cite{oquab2024dinov2} features capture the appearance of the object.
\method{} pools both with the object mask into object tokens and conditions a joint trajectory denoiser on them, so that the prior itself places the object in 3D, where the interaction takes place; removing these features lowers the ADD@.3 of the prior by \AbNoVfmHotGvPD{} points on HOT3D.
Because contact is part of the sampled state, this contact guides a short refinement that fits the object to the observed masks, so \method{} needs neither an object pose estimator nor an external contact model at inference.

Our contributions are as follows:
\begin{itemize}
\item We propose \method{}, a hand--object trajectory prior conditioned on frozen foundation features of egocentric video, which places the object in 3D by itself and refines the input hand track.
\item We show that the per-hand contact sampled by the prior can replace external contact labels in guiding a short refinement to the observed masks.
\item We demonstrate that \method{} achieves state-of-the-art object pose accuracy on HOT3D, where even the prior alone exceeds WHOLE, transfers to H2O without fine-tuning, and more than doubles the success rate of robot retargeting in a physics simulator over a dedicated object pose estimator.
\end{itemize}

\section{Related Work}
\label{sec:related}

\paragraph{Image-based 3D hand--object reconstruction.}
Single-view methods estimate MANO~\cite{romero2017mano} hand meshes~\cite{pavlakos2024hamer,potamias2025wilor} and the 6-DoF pose of novel objects from CAD models or reference views~\cite{labbe2022megapose,wen2024foundationpose,ornek2024foundpose}, but in a grasp the two occlude and constrain each other, so estimating them separately leaves their relative pose ambiguous.
Joint methods couple them through contact and penetration terms~\cite{hasson2019obman,grady2021contactopt}, implicit fields shared by hand and object~\cite{karunratanakul2020grasping,chen2022alignsdf}, hand-conditioned object prediction~\cite{ye2022ihoi,chen2025hort}, constraints in the wild~\cite{cao2021mow} or foundation-model priors~\cite{liu2025easyhoi}.
However, these methods, like recent egocentric estimators~\cite{bansal2026hopformer}, work one frame at a time, and their estimates chained over a video jitter.

\paragraph{Video-based 4D hand--object reconstruction.}
Video adds temporal evidence and, with camera tracking, a common world frame.
For hands, temporal attention~\cite{ye2026haptic}, world-space estimation~\cite{zhang2025hawor,yu2025dynhamr} and motion priors~\cite{zeng2026stablehand} recover hand motion over time, and V-JEPA Hand~\cite{li2026vjepahand} regresses both hands and the camera from egocentric video in one pass.
For objects, 6-DoF trackers follow a known or progressively reconstructed model, mostly in RGB-D video~\cite{wen2021bundletrack,wen2023bundlesdf,wen2024foundationpose}, the hand can guide object tracking in RGB~\cite{shin2026compose,zhu2026rohit}, and object trajectories have been recovered even from internet videos~\cite{ponimatkin2025object}.
Most of these methods model one side only, although in manipulation the hand both occludes the object and reveals how it moves.
Methods that model both pursue one of two goals.
Shape-centric methods reconstruct unknown objects from short interaction clips by per-sequence optimization~\cite{fan2024hold,on2025bigs}, generative priors~\cite{ye2023diffhoi,ye2024ghop,wang2025magichoi} or feed-forward prediction~\cite{chen2026forehoi}.
Motion-centric methods track hand and object poses over time under physical~\cite{hu2022physical} or contact~\cite{xu2026openchoir} constraints or with Gaussian models initialized from pretrained estimates~\cite{aytekin2026grag}, denoise hand motion against a given object trajectory~\cite{liu2024geneoh,luo2024physicshoi} or, concurrently, regress camera, hand and object poses jointly~\cite{ji2026egoxperience}.
Closest to ours are generative priors over hand--object trajectories, which learn how hands and objects move together and thus stay plausible under occlusion.
What limits them is how the object is placed in 3D.
WHOLE~\cite{ye2026whole} conditions its prior only on the template and a hand track, so the video locates the object only through 2D mask reprojection guidance at every sampling step, with VLM-derived contact labels.
DynamicHOI~\cite{guo2026dynamichoi} and 4D-HOF~\cite{li2026hof} instead refine the output of separate hand and object pose estimators~\cite{potamias2025wilor,wen2024foundationpose}, and the template-free EgoGrasp~\cite{fu2026egograsp} completes object poses tracked from foundation-model estimates.
Where 3D geometry is used, it comes from a separate depth estimator as a target to fit~\cite{li2026hof,fu2026egograsp,aytekin2026grag}.
\method{} instead makes frozen 3D geometry~\cite{wang2025pi3} and appearance~\cite{oquab2024dinov2} features of the video a condition of its learned prior, in the setting of WHOLE; masks select the object and enter a short refinement, where sampled contact replaces VLM-derived labels.

\paragraph{From human video to robot manipulation.}
Human video is an abundant source of dexterous demonstrations, and reconstructed hand motion~\cite{shaw2022videodex}, or hand and object motion~\cite{qin2022dexmv}, is the common interface to it.
Imitating hand joints alone, however, does not reproduce the interaction~\cite{park2025learning}; whether the object follows depends on its own trajectory and on when the hand holds it.
Transfer methods track reference motions of both the hand and the object in simulation, as ManipTrans~\cite{li2025maniptrans} and others~\cite{liu2024quasisim,liu2025dextrack,zhu2026chord} do with motion capture or curated in-house videos.
Do as I Do~\cite{paliwal2026doasido} and OmniHOI~\cite{mao2026omnihoi} instead bring monocular human videos to robots by reconstructing hands and objects with vision models, as do data engines~\cite{zhang2025robowheel,wang2026egoinfinity} and video-to-policy methods~\cite{chen2026manipulation,chen2026v2pmanip}, while object-trajectory transfer~\cite{ponimatkin2025object} keeps only the object.
\method{} can supply such pipelines with hand and object references from egocentric video, which \cref{sec:exp:robot} tests in physics-based retargeting.

\section{Method}
\label{sec:method}

\begin{figure*}[t]
\centering
\includegraphics[width=\linewidth]{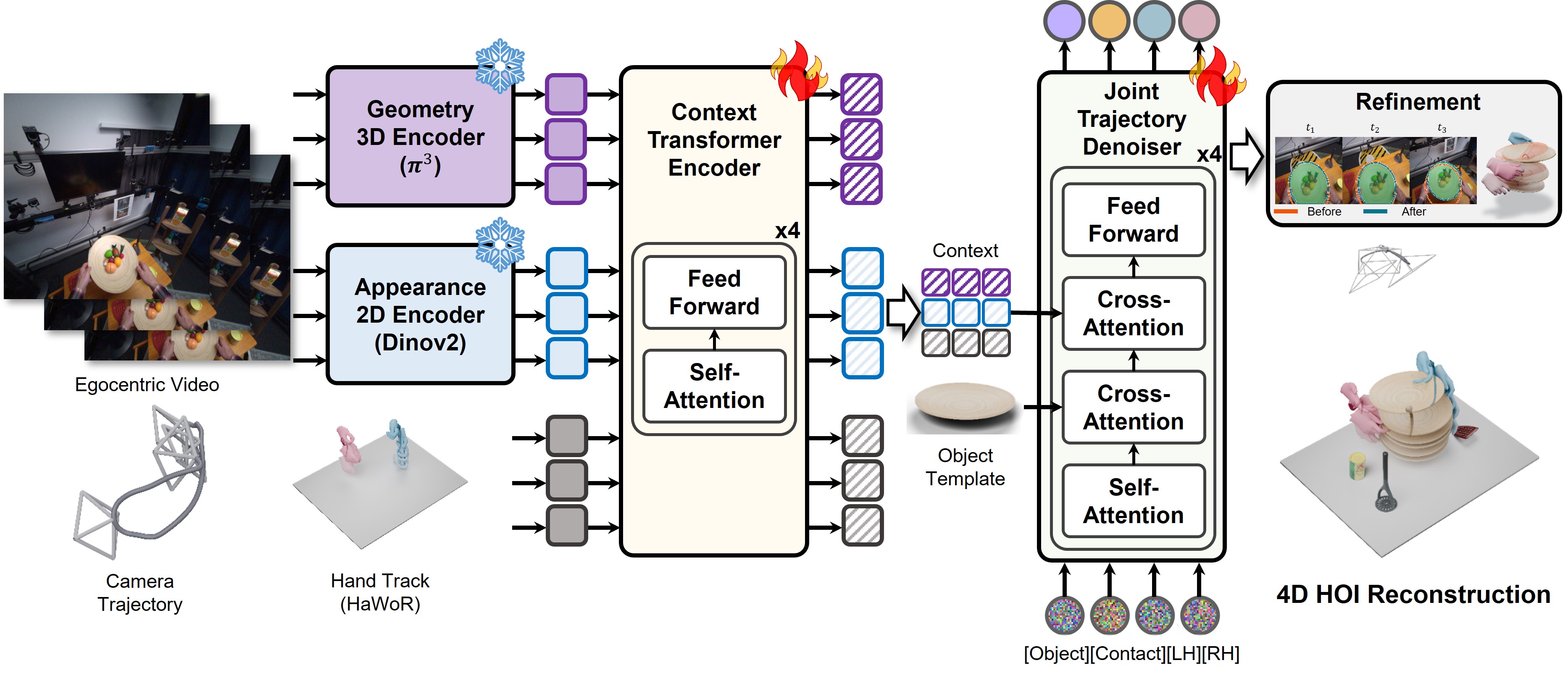}
\caption{\textbf{Overview of \method{}.} A context Transformer encoder mixes frozen $\pi^3$ (3D geometry) and DINOv2 (2D appearance) features, their mask-pooled object tokens, and tokens of the hand track $\bar{H}$ and camera trajectory into the context $C$. Attending to $C$ and to the template tokens $S$, the joint trajectory denoiser $D_\phi$ denoises one token $x_t$ per frame with the object pose, the per-hand contact and both hands (LH, RH). The refinement then fits the object to the observed masks $M_t$ (orange: before, teal: after). Snowflake: frozen; flame: trained.}
\label{fig:overview}
\end{figure*}

We reconstruct the MANO motion of both hands, the 6-DoF object trajectory and per-hand contact by using an off-the-shelf hand track as a reference and placing the object in 3D around the hands, which requires the 3D layout of the scene and the pixels that belong to the object.
Frozen visual foundation models (VFMs) provide these cues, 3D geometry from $\pi^3$~\cite{wang2025pi3} and appearance from DINOv2~\cite{oquab2024dinov2}, and \method{} conditions its prior directly on them.
In stage 1, the \emph{prior}, a video-conditioned diffusion model, jointly samples all three quantities for a window of frames (\cref{fig:overview,sec:method:prior,sec:method:training}).
In stage 2, the \emph{refinement} fits the sampled object trajectory to the masks once, guided by the sampled contact (\cref{sec:method:refine}).

\subsection{Problem setup and representation}
\label{sec:method:setup}
We process windows of $T$ frames.
The inputs are the RGB frames $\{I_t\}$, the camera poses from the headset's SLAM, the object template mesh $\mathcal{O}$ and its visible masks $\{M_t\}$.
A coarse world-space hand track $\bar{H}$ comes from an off-the-shelf estimator~\cite{zhang2025hawor}, and no object pose estimate is given.
All poses are expressed in the frame of the first camera of the window.
Following WHOLE, the state of each frame concatenates the object pose, the contact of each hand and the MANO parameters of each hand.
Specifically, $x_t\in\mathbb{R}^{73}$ holds the object rotation (6D~\cite{zhou2019continuity}) and translation, a contact value $c^h_t$ for each hand $h\in\{\mathrm{L},\mathrm{R}\}$ and, for each hand, the global rotation $\Gamma^h_t$, the wrist translation $\Lambda^h_t$, 15 PCA coefficients $\Theta^h_t$ of the finger articulation and the 10 shape coefficients $\beta^h$.
A window is $x=(x_1,\dots,x_T)$.
The ground-truth contact of a hand is 1 when any vertex of its mesh lies within 5\,mm of the object surface and 0 otherwise, and at inference we threshold the sampled value at 0.5.
To refine the hand track rather than replace it, we predict the global pose of each hand as a residual on the track,
\begin{equation}
R(\Gamma^h_t)=\operatorname{Exp}(\Delta\Gamma^h_t)\,R(\bar{\Gamma}^h_t),\quad \Lambda^h_t=\bar{\Lambda}^h_t+\Delta\Lambda^h_t,
\label{eq:residual}
\end{equation}
where $R(\cdot)$ maps an axis-angle vector to a rotation matrix.

\subsection{Video-conditioned hand--object prior}
\label{sec:method:prior}
\paragraph{VFM context encoder.}
Each frame is encoded by two frozen VFMs (\cref{fig:overview}).
The 3D geometry encoder $\pi^3$~\cite{wang2025pi3} gives the decoder features of its last two blocks, computed with attention across frames in chunks of 16, and the 2D appearance encoder DINOv2~\cite{oquab2024dinov2} gives patch tokens of single-view appearance.
A Perceiver resampler~\cite{alayrac2022flamingo} compresses each patch map into a few tokens per frame.
Pooling each patch map with the soft object mask adds one geometry and one appearance token for the object; for an empty mask, the geometry and appearance tokens reduce to a learned constant and to zero, respectively.
The hand track $\bar{H}_t$ and the Fourier-encoded camera motion relative to the first frame each add one token per frame.
A context Transformer encoder mixes all tokens, with type embeddings and temporal positional encodings, into the context $C$.
The template is encoded as a basis point set~\cite{prokudin2019bps} and resampled into template tokens $S$.

\paragraph{Joint trajectory denoiser.}
The denoiser $D_\phi(x^{(n)}, n, C, S)$ takes the noisy window $x^{(n)}$ at diffusion step $n$ as one token per frame that concatenates the object pose, the contact and both hands.
Each block applies self-attention across frames, cross-attention to $S$ and to $C$, and a feed-forward layer, with adaptive layer normalization~\cite{peebles2023dit} conditioned on the diffusion step, the pooled template and a per-frame mask reliability.
This reliability, the visible mask area relative to the largest in the window, lets the denoiser discount frames where the object is barely visible.
The denoiser predicts the clean window $\hat{x}_0$, and we use standard DDPM sampling~\cite{ho2020ddpm}.

\subsection{Training}
\label{sec:method:training}
For diffusion, $x_0$ denotes the state with the hand global poses represented as residuals on $\bar{H}$ and all channels normalized by their training-set statistics.
The main objective is the denoising loss
\begin{equation}
\mathcal{L}_\mathrm{diff}=\mathbb{E}_{x_0,n,\epsilon}\Big[\frac{1}{T}\sum_{t}\frac{\sum_j w_j\big(\hat{x}_{0,tj}-x_{0,tj}\big)^2}{\sum_j w_j}\Big],
\label{eq:diff}
\end{equation}
where $x^{(n)}$ is $x_0$ noised to step $n$ with Gaussian noise $\epsilon\sim\mathcal{N}(0,I)$ by the forward process, $\hat{x}_0=D_\phi(x^{(n)},n,C,S)$ is the denoiser output and $j$ indexes the state channels.
The channel weights $w_j$ emphasize the object and contact channels over the hand and shape channels.
Rather than training a separate contact classifier, we learn contact as part of the state through this loss.

A hand holding the object indicates where the object is, and our contact-conditioned relative-pose loss turns the interaction loss of WHOLE~\cite{ye2026whole} into a supervised one.
For each hand we build a wrist transform $F^h_t\in SE(3)$ from three MANO joints, with its origin at the wrist, and express the object pose $T^o_t$ in this frame, $X^h_t=(F^h_t)^{-1}T^o_t$.
On the set $\mathcal{C}$ of frames and hands with ground-truth contact $c^{h\star}_t=1$, the predicted relative pose is supervised by the ground-truth one,
\begin{equation}
\mathcal{L}_\mathrm{rel}=\frac{1}{|\mathcal{C}|}\textstyle\sum_{(h,t)\in\mathcal{C}}\Big(\big\|\hat{\tau}^{\,h}_t-\tau^h_t\big\|_1+\big\|\hat{R}^h_t-R^h_t\big\|_F^2\Big),
\label{eq:rel}
\end{equation}
where $(R^h_t,\tau^h_t)$ are the rotation and translation of $X^h_t$.
Its gradient reaches both the object pose and, through MANO forward kinematics, the hand pose.
The total loss is $\mathcal{L}=\mathcal{L}_\mathrm{diff}+\lambda_\mathrm{rel}\mathcal{L}_\mathrm{rel}$ with $\lambda_\mathrm{rel}=4.1\times10^{-3}$, ramped in after the denoiser has started to converge (Supp.~\cref{tab:supp_hparams}).  %
To make the prior robust to missing masks, we drop the mask for the whole window or for random contiguous segments during training, and we drop the template with a small probability.

\subsection{Refinement after sampling}
\label{sec:method:refine}
From one sample of the prior, we optimize only the object rotations and translations $\{(R_t,\tau_t)\}$ for a fixed number of steps while keeping the hands fixed.
The mask (silhouette) term $E_\mathrm{sil}$ projects points sampled on the template surface into each frame and penalizes their distance to the mask in both directions: projected points outside the mask by the mask's distance transform, and points on the mask boundary by the distance to the nearest projected point.
A smoothness term $E_\mathrm{smooth}$ penalizes the acceleration of the object center, and a prior term $E_\mathrm{prior}=\sum_t\|\tau_t-\hat{\tau}_t\|^2$ keeps the translation near the sample $\hat{\tau}$.
The sampled contact, rather than an external contact model~\cite{ye2026whole}, guides the refinement through a static term.
Whenever the sampled contact of both hands stays below the threshold for at least two consecutive frames, the object is treated as resting.
Its center $p_t=R_t p^{\mathcal{O}}+\tau_t$, with $p^{\mathcal{O}}$ the template center, is then pulled to the segment mean:
\begin{equation}
E_\mathrm{static}=\frac{1}{T}\textstyle\sum_{s}\sum_{t\in s}\|p_t-\bar{p}_s\|^2,
\label{eq:static}
\end{equation}
where $s$ ranges over such non-contact segments and $\bar{p}_s$ is the mean center over $s$.
For rotation, we use a smoothness term on the second difference of relative rotations and a prior on the rotation update, and the refinement minimizes the weighted sum of all terms.

\subsection{Implementation details}
\label{sec:method:impl}
The prior operates on windows of $T=120$ frames, so we sample and refine each 150-frame clip as two windows with a stride of 30 and transform their hand and object poses into a common world frame.
Overlapping windows are merged by a crossfade and temporal smoothing, keeping the contact of the more heavily weighted window (Supp.~\cref{sec:supp_impl}).
The context encoder and the denoiser each have 4 layers of width 512 with 8 heads.
Images are resized to $224\times224$, giving 256 patches per frame, and the features of each backbone are resampled to 16 tokens per frame.
We train from scratch with AdamW at a learning rate of $2\times10^{-4}$, batch size 8, for 100k iterations, and use an exponential moving average of the weights for evaluation.
Training takes about 60 hours on a single RTX 5090.  %
The diffusion uses 1000 steps with a linear noise schedule, and we draw one sample per window.
Refinement runs 200 Adam steps on the translation alone, with the rotation fixed to the sample, and then 200 steps on the full pose.
The prior has 68.4M trainable parameters; the frozen $\pi^3$ and DINOv2 backbones are not fine-tuned.

\section{Experiments}
\label{sec:experiments}
\begin{table*}[t]
\centering
\caption{\textbf{Hand and object reconstruction on HOT3D.} W and WA in cm, PA in mm. \method{} uses ground-truth visible masks, and ``(prior)'' denotes the sample of our prior before refinement. $^\dagger$Evaluated by us on our 147 clips with the same inputs as ours; the HaWoR row is its native output, not our converted input track. $^*$Taken from the original papers, which use their own test clips; FoundationPose as reproduced by DynamicHOI is in Supp.~\cref{tab:supp_hot3d}. Best per column across both groups: object in bold, hand underlined. Ours shaded.}  %
\label{tab:hot3d}
\small
\begin{tabular}{lcccccc}
\toprule
 & \multicolumn{3}{c}{Object} & \multicolumn{3}{c}{Hand} \\
\cmidrule(lr){2-4}\cmidrule(lr){5-7}
Method & ADD@.1$\uparrow$ & ADD@.3$\uparrow$ & ADD-S@.3$\uparrow$ & W$\downarrow$ & WA$\downarrow$ & PA$\downarrow$ \\
\midrule
\multicolumn{7}{l}{\textit{Evaluated on our 147 clips}} \\
HaWoR$^\dagger$~\cite{zhang2025hawor} & -- & -- & -- & \HotHaworW & \HotHaworWA & \HotHaworPA \\
WiLoR$^\dagger$~\cite{potamias2025wilor} & -- & -- & -- & \HotWilorW & \HotWilorWA & \HotWilorPA \\
FoundationPose$^\dagger$~\cite{wen2024foundationpose} & \FpOursAddA & \FpOursAddC & \FpOursAddsC & -- & -- & -- \\
\oursrow \method{} (prior) & \HotGvPriorAddA & \HotGvPriorAddC & \HotGvPriorAddsC & \HotGvW & \HotGvWA & \HotGvPA \\
\oursrow \method{} & \textbf{\HotGvAddA} & \textbf{\HotGvAddC} & \textbf{\HotGvAddsC} & \HotGvW & \HotGvWA & \HotGvPA \\
\midrule
\multicolumn{7}{l}{\textit{Results from the original papers}} \\
WHOLE$^*$~\cite{ye2026whole} & -- & \WholeAddC & \WholeAddsC & \WholeW & \WholeWA & \WholePA \\
DynamicHOI$^*$~\cite{guo2026dynamichoi} & \DynAddA & \DynAddC & \DynAddsC & \underline{\DynW} & \underline{\DynWA} & \underline{\DynPA} \\
\bottomrule
\end{tabular}
\end{table*}

\subsection{Experimental setup}
\label{sec:exp:setup}
\paragraph{Datasets.}
We train and evaluate on HOT3D~\cite{banerjee2025hot3d}, egocentric Aria recordings of two-handed manipulation with ground-truth hands, object poses and meshes.
Matching the split sizes of DynamicHOI~\cite{guo2026dynamichoi}, we train on 1,311 clips and test on 147 clips of 150 frames.
Following the HOT3D split of the hand estimator HaWoR~\cite{zhang2025hawor}, the test clips come from 10 of its validation recordings, which neither model sees in training.
To test generalization, we apply the model without fine-tuning to the 217 clips of the official H2O~\cite{kwon2021h2o} test split with ground-truth object poses, whose objects are unseen in HOT3D.

\paragraph{Inputs.}
Following WHOLE~\cite{ye2026whole}, all object reconstruction methods are given the object template and the camera trajectory from the headset's SLAM (on H2O, the camera poses of the dataset).
The hand track is derived from HaWoR~\cite{zhang2025hawor} estimates (Supp.~\cref{sec:supp_inputs}).
Our main results use ground-truth visible object masks, and \cref{sec:exp:analysis} uses masks predicted by \samthree{}~\cite{carion2026sam3}; the same masks serve as model input and as refinement targets.
We evaluate every object that a hand touches in a clip (228 tracks), identified from the ground-truth contact.
WHOLE instead evaluates 50 held-out object trajectories that move by more than 5\,cm; on the \DynSubN{} of our tracks that satisfy this rule, \method{} reaches \DynSubGvAddC{} ADD@.3.  %

\paragraph{Evaluation metrics.}
For hands, we report the world-frame MPJPE after aligning the first two frames (W) and the whole sequence (WA), in cm, and the Procrustes-aligned MPJPE (PA), in mm.
For objects, we report the area under the accuracy curve of ADD and of the symmetry-aware ADD-S up to 0.1\,m and 0.3\,m (ADD@.1, ADD@.3, ADD-S@.3).  %
In the text, values in brackets denote 95\% bootstrap confidence intervals of paired differences (Supp.~\cref{sec:supp_eval}).  %

\paragraph{Baselines.}
We compare with WHOLE~\cite{ye2026whole} and DynamicHOI~\cite{guo2026dynamichoi}, the most recent template-based methods evaluated on HOT3D.
As their official code is not available, we use the numbers reported in their papers on their own test clips.
We also run FoundationPose~\cite{wen2024foundationpose} on our clips with the same masks, template and camera as ours, registering it on the first visible frame and tracking with metric depth from Depth Anything~3~\cite{lin2025depthanything3}.
We further evaluate the hand estimators HaWoR~\cite{zhang2025hawor} and WiLoR~\cite{potamias2025wilor}.

\subsection{Comparison on HOT3D}
\label{sec:exp:hot3d}

\begin{table}[t]
\centering
\caption{\textbf{Object ADD@.3 on the frame subsets of WHOLE} on HOT3D. Truncated: the object is partly outside the image; out of view: entirely outside, split by our implementation. HOI: object error after aligning the hand trajectory to the ground truth. $^*$Taken from the original paper. Best in bold.}
\label{tab:subsets}
\footnotesize
\setlength{\tabcolsep}{3.5pt}
\begin{tabular}{lccccc}
\toprule
Method & All & Contact & Trunc. & Out of view & HOI \\
\midrule
WHOLE$^*$~\cite{ye2026whole} & \WholeAddC & \WholeContactAdd & \WholeTruncAdd & \WholeOovAdd & \WholeHoiAdd \\
\oursrow \method{} (prior) & \HotGvPriorAddC & \SubGvPriorContactAdd & \SubGvPriorTruncAdd & \SubGvPriorOovAdd & \HoiGvPriorAdd \\
\oursrow \method{} & \textbf{\HotGvAddC} & \textbf{\SubGvContactAdd} & \textbf{\SubGvTruncAdd} & \textbf{\SubGvOovAdd} & \textbf{\HoiGvAdd} \\
\bottomrule
\end{tabular}
\end{table}

\begin{figure*}[t]
\centering
\includegraphics[width=0.9\linewidth]{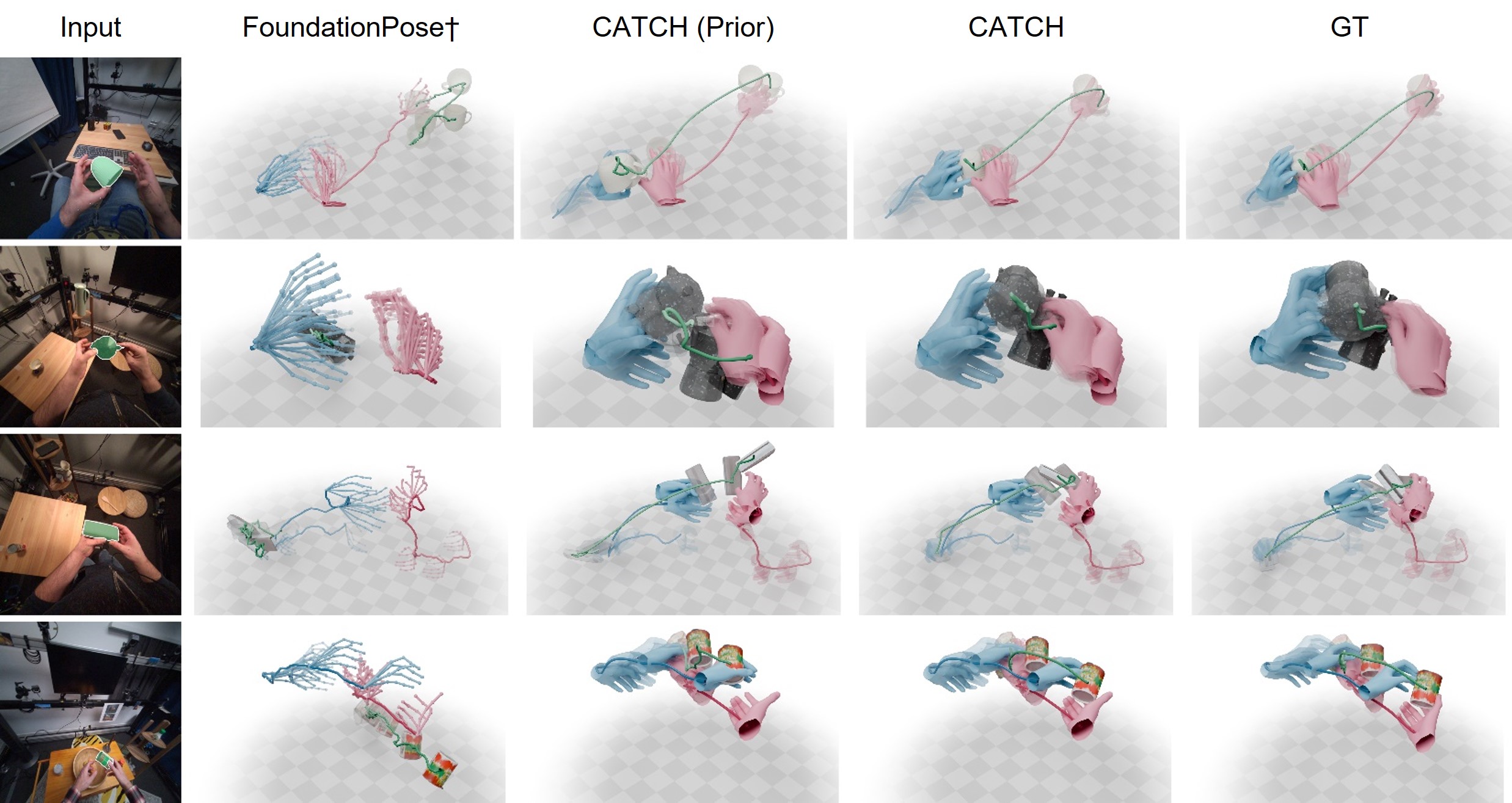}
\caption{\textbf{Qualitative comparison on HOT3D.} Each row shows a test clip, with an input frame and its object mask on the left. FoundationPose$^\dagger$ is shown with the HaWoR hand track as skeletons. ``\method{} (prior)'' shows a sample from our prior, and ``\method{}'' shows the result after refinement. The two hands are shown in blue and pink with their wrist trajectories, and the object trajectory in green; earlier frames are more transparent.}
\label{fig:qual}
\end{figure*}

Because the prior itself places the object, \method{} achieves the highest object accuracy among the compared methods (\cref{tab:hot3d}).
With the same inputs on the same clips, FoundationPose reaches less than half of our ADD@.3 (\FpOursAddC{} vs \HotGvAddC{}), and even our prior alone (\HotGvPriorAddC{}) doubles it.
In \cref{fig:qual}, FoundationPose places the object far from the hands, whereas \method{} keeps it in the holding hand.
The margin over WHOLE is largest out of view, where the mask is empty and 2D alignment has nothing to fit (\cref{tab:subsets}).
Even there our prior alone exceeds WHOLE (\SubGvPriorOovAdd{} vs \WholeOovAdd{}), and the refinement then reaches \SubGvOovAdd{} as its smoothness term carries the fit from neighboring masked frames.
As the refinement keeps the hands fixed, the prior sets hand accuracy and lowers the WA and PA errors of HaWoR, from which its input hand track is derived.

\subsection{Zero-shot generalization to H2O}
\label{sec:exp:h2o}

\begin{table}[tb]
\centering
\caption{\textbf{Zero-shot object reconstruction on H2O.} \method{} uses ground-truth visible masks, rendered from the annotated poses. $^*$Reported in WHOLE~\cite{ye2026whole}. $^\ddagger$Trained on H2O; all other methods are trained on HOT3D only. Best in bold.}
\label{tab:h2o}
\footnotesize
\setlength{\tabcolsep}{4pt}
\begin{tabular}{lcc}
\toprule
Method & ADD@.3$\uparrow$ & ADD-S@.3$\uparrow$ \\
\midrule
H2OTR$^{*\ddagger}$~\cite{cho2023transformer} & \HtotrHtoAdd & \HtotrHtoAdds \\
WHOLE$^*$~\cite{ye2026whole} & \WholeHtoAdd & \WholeHtoAdds \\
\midrule
\oursrow \method{} (prior) & \HtoGvPriorAddC{} & \HtoGvPriorAddsC{} \\
\oursrow \method{} & \textbf{\HtoGvAddC} & \textbf{\HtoGvAddsC} \\
\bottomrule
\end{tabular}
\end{table}

\begin{figure}[tb]
\centering
\includegraphics[width=0.65\linewidth]{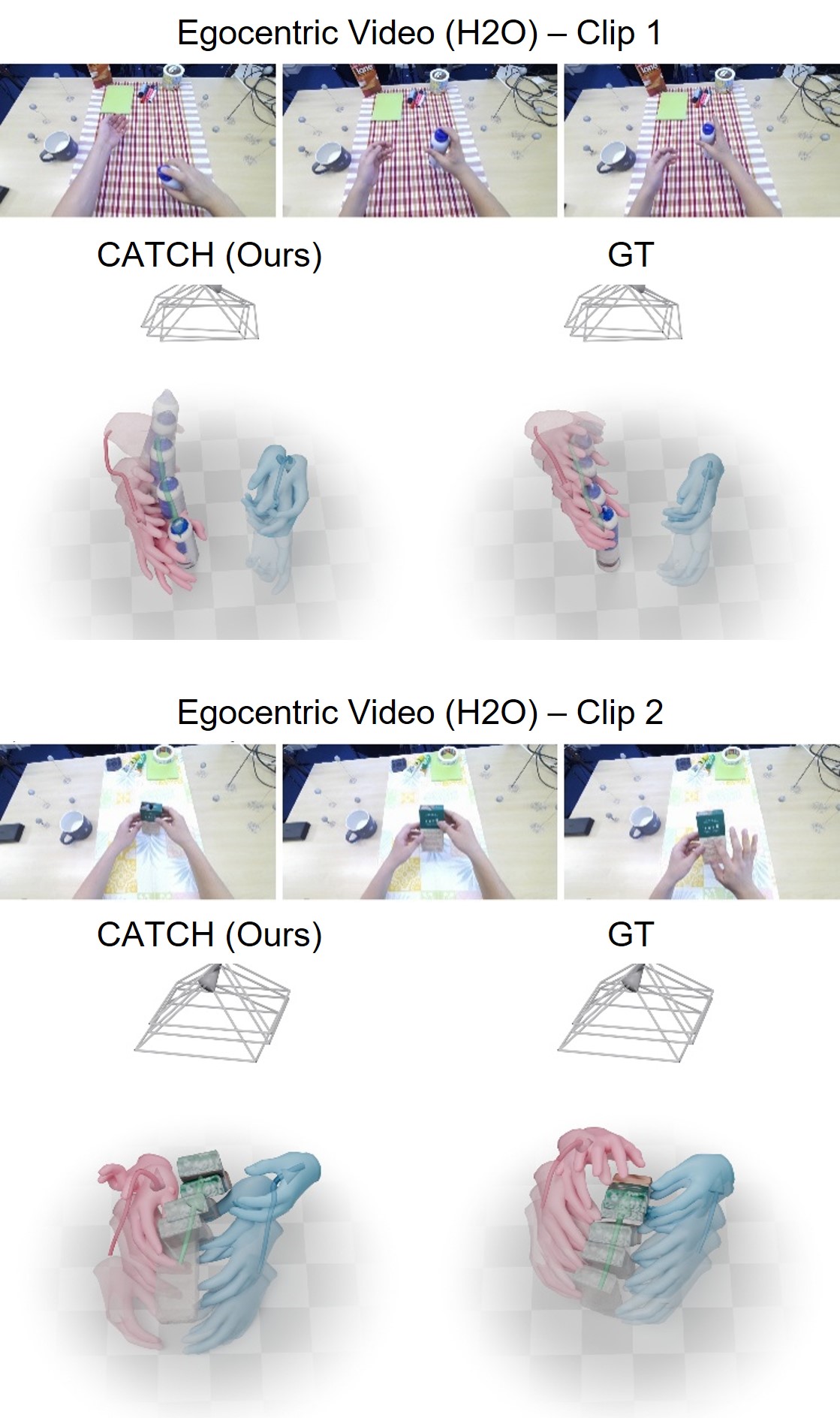}
\caption{\textbf{Zero-shot reconstruction on H2O} by \method{}, trained only on HOT3D. For two test clips with objects unseen in training, we show input frames (top) and our reconstruction next to the ground truth (bottom), with the camera frustum. Ground-truth visible masks are used as input.}
\label{fig:h2o}
\end{figure}

Trained only on HOT3D and refined with the same weights, \method{} outperforms H2OTR, which is trained on H2O, in both metrics, with an ADD-S@.3 of \HtoGvAddsC{} against \HtotrHtoAdds{} (\cref{tab:h2o,fig:h2o}).
Its ADD@.3 of \HtoGvAddC{} is also well above the \WholeHtoAdd{} that WHOLE reaches without fine-tuning.

\subsection{Analysis}
\label{sec:exp:analysis}

\begin{table}[tb]
\centering
\caption{\textbf{Effect of visual conditioning on HOT3D.} Separate models trained on the same 1,311 clips with the same seed and evaluated with ground-truth visible masks: ADD@.3 of the prior, and ADD@.1 and ADD@.3 after the same refinement. Removing both feature sets also removes the mask-pooled object tokens, the only input that carries the mask location. Results with \samthree{} masks are in Supp.~\cref{tab:supp_ablation}. Best in bold.}
\label{tab:ablation}
\footnotesize
\setlength{\tabcolsep}{4pt}
\begin{tabular}{lccc}
\toprule
 & Prior & \multicolumn{2}{c}{Refinement} \\
\cmidrule(lr){2-2}\cmidrule(lr){3-4}
Model & ADD@.3$\uparrow$ & ADD@.1$\uparrow$ & ADD@.3$\uparrow$ \\
\midrule
\method{} & \textbf{\AbFullHotGvP} & \textbf{\AbFullHotGvRA} & \textbf{\AbFullHotGvR} \\
w/o $\pi^3$ & \AbNoPiHotGvP & \AbNoPiHotGvRA & \AbNoPiHotGvR \\
w/o DINOv2 & \AbNoDinoHotGvP & \AbNoDinoHotGvRA & \AbNoDinoHotGvR \\
w/o $\pi^3$, DINOv2 & \AbNoVfmHotGvP & \AbNoVfmHotGvRA & \AbNoVfmHotGvR \\
\bottomrule
\end{tabular}
\end{table}

\paragraph{Effect of visual conditioning.}
To test whether the features place the object, we train separate models without $\pi^3$, without DINOv2 and without both (\cref{tab:ablation}).
With the same masks and the same refinement, removing both feature sets lowers ADD@.1 from \AbFullHotGvRA{} to \AbNoVfmHotGvRA{} (\AbNoVfmHotGvRAD{} [\AbNoVfmHotGvRALo{}, \AbNoVfmHotGvRAHi{}]), and the prior loses \AbNoVfmHotGvPD{} ADD@.3 points.
Each feature set contributes on its own, and DINOv2 also reduces the W error of the hands by \AbNoDinoHotGvHandD{}~cm, so the prior also refines the hands from video.

\begin{table}[tb]
\centering
\caption{\textbf{Object reconstruction with masks predicted by \samthree{}.} The predicted masks serve as model input and as refinement targets; the last row lists our results with ground-truth visible masks from \cref{tab:hot3d,tab:h2o}. $^*$Reported by the authors. Best in bold, second best underlined.}
\label{tab:masks}
\footnotesize
\setlength{\tabcolsep}{3pt}
\resizebox{\linewidth}{!}{%
\begin{tabular}{@{}lcccccc@{}}
\toprule
 & \multicolumn{4}{c}{HOT3D} & \multicolumn{2}{c}{H2O} \\
\cmidrule(lr){2-5}\cmidrule(lr){6-7}
Method & ADD@.1$\uparrow$ & ADD-S@.1$\uparrow$ & ADD@.3$\uparrow$ & ADD-S@.3$\uparrow$ & ADD@.3$\uparrow$ & ADD-S@.3$\uparrow$ \\
\midrule
WHOLE$^*$~\cite{ye2026whole} & -- & -- & \WholeAddC & \WholeAddsC & \WholeHtoAdd & \WholeHtoAdds \\
DynamicHOI$^*$~\cite{guo2026dynamichoi} & \DynAddA & \DynAddsA & \underline{\DynAddC} & \underline{\DynAddsC} & -- & -- \\
\midrule
\oursrow \method{} (prior) & \HotSmPriorAddA & \HotSmPriorAddsA & \HotSmPriorAddC & \HotSmPriorAddsC & \HtoSmPriorAddC & \HtoSmPriorAddsC \\
\oursrow \method{} & \underline{\HotSmAddA} & \underline{\HotSmAddsA} & \HotSmAddC & \HotSmAddsC & \underline{\HtoSmAddC} & \underline{\HtoSmAddsC} \\
\midrule
\oursrow \method{} (GT visible) & \textbf{\HotGvAddA} & \textbf{\HotGvAddsA} & \textbf{\HotGvAddC} & \textbf{\HotGvAddsC} & \textbf{\HtoGvAddC} & \textbf{\HtoGvAddsC} \\
\bottomrule
\end{tabular}}
\end{table}

\paragraph{Predicted masks.}
In practice, masks come from a segmentation model; \method{} also works with masks predicted by \samthree{} (\cref{tab:masks}), and its values remain above those reported by WHOLE on both datasets.
Most of the drop comes from detection failures: on HOT3D, \SamFailMissPct{}\% of the tracks that lose at least 5 ADD@.3 points have an empty or wrong mask, whereas tracks with accurate masks (mean IoU $\geq$ 0.8) lose only \SamGoodDropMax{} points on average (Supp.~\cref{fig:supp_sam_failure}).  %
Compared with the numbers reported by DynamicHOI, \method{} is ahead at the strict 0.1\,m threshold in both ADD and ADD-S.  %

\begin{figure}[tb]
\centering
\includegraphics[width=0.85\linewidth]{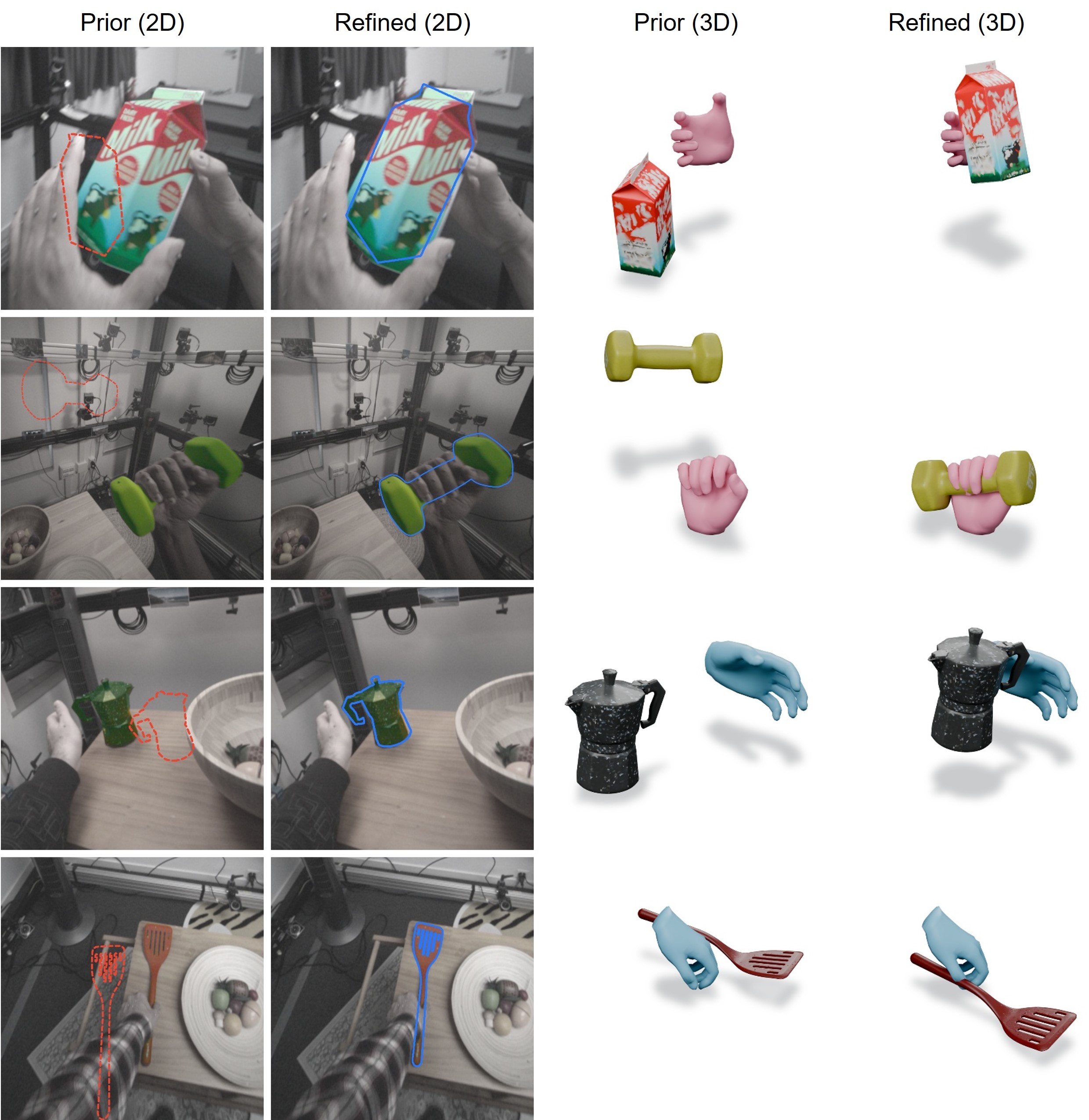}
\caption{\textbf{Effect of the refinement on HOT3D.} 2D: the object sampled by the prior (red dashed) and after refinement (blue), projected onto the input frame with the visible mask in green. 3D: the same object with the hand that holds it, before and after refinement.}
\label{fig:refine}
\end{figure}

\paragraph{Initialization and refinement.}
To test whether the refinement alone explains the accuracy, we apply it to a training-free PnP initialization of the object instead (on 1,000 held-out windows with a separate model; Supp.~\cref{sec:supp_analysis}).
It then ends \TestGvOursVsPnp{} ADD@.3 points below \method{}, and refining once after sampling matches or outperforms guidance inside the sampling loop (\cref{fig:refine}, Supp.~\cref{tab:supp_analysis}).

\paragraph{Contact in the refinement.}
We next test whether the sampled contact can replace ground-truth contact in the refinement.
On the 147 HOT3D clips, it agrees with the ground truth at an F1 score of \ConFOneGv{}\%, and using ground-truth contact instead changes ADD@.3 by only \ConGtGvD{} points, so no contact model is needed.  %
Removing the static term or treating the object as always resting lowers ADD@.3, also with \samthree{} masks (Supp.~\cref{sec:supp_sam}).

\subsection{Robot retargeting}
\label{sec:exp:robot}
\begin{table}[tb]
\centering
\caption{\textbf{Physics-based retargeting to a robot hand.} Success rate (\%) of SPIDER~\cite{pan2026spider} rollouts over 51 HOT3D tracks with three seeds each, for each pair of hand and object references. A rollout succeeds when the final object position lies within 10\,cm of the ground truth. Bold: best reconstructed object with ground-truth hands.}
\label{tab:robot}
\footnotesize
\begin{tabular}{llc}
\toprule
Hand & Object & Success$\uparrow$ \\
\midrule
GT & GT & \RobGtGt \\
\midrule
GT & FoundationPose~\cite{wen2024foundationpose} & \RobGtFp \\
GT & \method{} (prior) & \RobGtPrior \\
GT & \method{} & \textbf{\RobGtOurs} \\
\midrule
\method{} & GT & \RobOursGt \\
\method{} & \method{} & \RobOursOurs \\
\bottomrule
\end{tabular}
\end{table}
\begin{figure}[tb]
\centering
\includegraphics[width=0.85\linewidth]{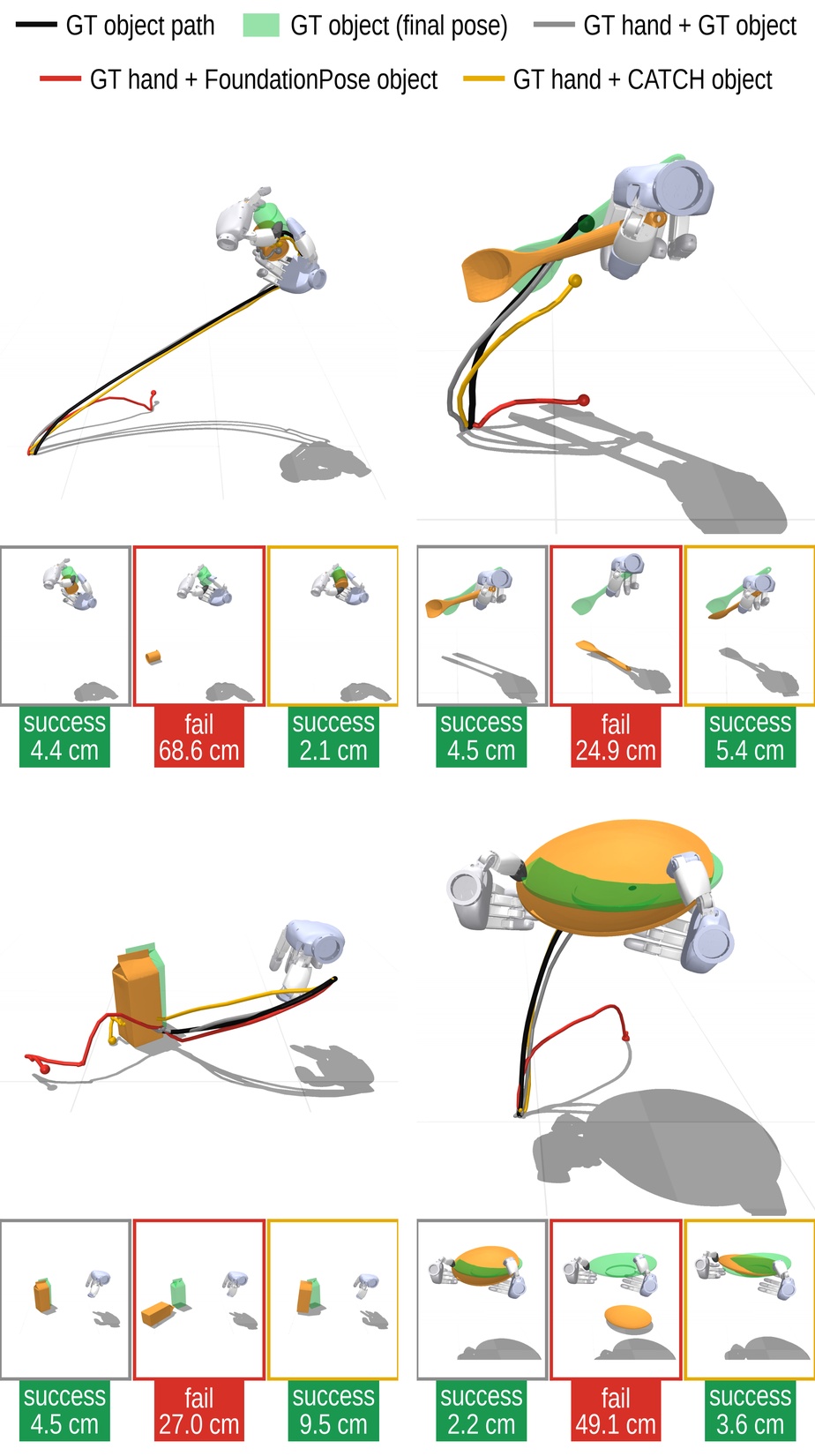}
\caption{\textbf{Robot retargeting rollouts} for four HOT3D tracks (seed 0). Simulated object trajectories when retargeting the ground-truth hand with the ground-truth object (gray), with the FoundationPose object (red) and with the object of \method{} (yellow); black: ground-truth object trajectory. Insets: final state (simulated object in orange, final ground-truth pose in green) and object position error; a rollout succeeds below 10\,cm.}
\label{fig:robot}
\end{figure}

Beyond pose metrics, we test whether the reconstructions can teach a robot hand through physics-based retargeting, using SPIDER~\cite{pan2026spider} to optimize the controls of a simulated XHAND hand so that the object follows the reference.
To exclude infeasible retargeting, we use the 51 hand--object tracks of our HOT3D clips on which retargeting the ground truth succeeds in at least two of three seeds.\footnote{In all rows, the simulation starts from the ground-truth first frame, and the wrist orientation of the input hand is aligned by a constant rotation offset per clip and hand, estimated from the ground-truth hand.}
To attribute the success, we keep either the hand or the object at the ground truth and replace the other with a reconstruction.

The object accuracy of \method{} carries over to the final object position in simulation (\cref{tab:robot,fig:robot}): with ground-truth hands, its objects succeed in \RobGtOurs{}\% of the rollouts, more than twice as often as those of FoundationPose (\RobGtFp{}\%), and even samples of the prior reach \RobGtPrior{}\%.

\section{Conclusion}
\label{sec:conclusion}
We presented \method{}, a diffusion prior that uses an off-the-shelf hand track as a reference and places the object in 3D from frozen foundation features of egocentric video, sampling both hands, the object trajectory and per-hand contact together.
The sampled contact then guides a short refinement to the masks.
\method{} achieves the highest object accuracy on HOT3D with ground-truth visible masks and transfers to H2O without fine-tuning.
In physics-based retargeting, its object trajectories let a simulated robot hand succeed far more often than those of a dedicated object pose estimator.
\paragraph{Limitations and future work.}
As in WHOLE~\cite{ye2026whole}, \method{} assumes a known object template and camera trajectory.
Next, we will extend the prior to unknown objects and in-the-wild video, and use its reconstructions to learn robot manipulation from egocentric human video.
Since the hand track enters as a condition, \method{} can adopt stronger hand trackers as they appear~\cite{li2026vjepahand}.

\section*{Acknowledgments}
This work was supported in part by the Industrial Strategic Technology Development Program (RS-2024-00442029) funded by the Ministry of Trade, Industry and Energy (MOTIE, Korea), in part by the National Research Foundation of Korea (NRF) funded by the Korea government (MSIT) (RS-2024-00464386), and in part by the KIST Institutional Program (26E0051 and 26Z9020).

{
    \small
    \bibliographystyle{ieeenat_fullname}
    \bibliography{ref}
}

\input{sec/X_suppl_arxiv}  %

\end{document}

%% file: numbers.tex
\newcommand{\HotGvAddA}{42.73}  %
\newcommand{\HotGvAddC}{78.47}  %
\newcommand{\HotGvAddsC}{91.43}  %
\newcommand{\HotSmAddA}{29.78}  %
\newcommand{\HotSmAddC}{64.58}  %
\newcommand{\HotSmAddsC}{77.60}  %
\newcommand{\HotGvPriorAddA}{19.03}  %
\newcommand{\HotGvPriorAddC}{64.95}  %
\newcommand{\HotGvPriorAddsC}{80.17}  %
\newcommand{\HotSmPriorAddA}{14.45}  %
\newcommand{\HotSmPriorAddC}{55.76}  %
\newcommand{\HotSmPriorAddsC}{70.19}  %
\newcommand{\HotGvW}{12.83}  %
\newcommand{\HotGvWA}{4.37}  %
\newcommand{\HotGvPA}{7.05}  %
\newcommand{\HotSmW}{12.99}  %
\newcommand{\HotSmWA}{4.36}  %
\newcommand{\HotSmPA}{7.02}  %
\newcommand{\HotHaworW}{12.92}  %
\newcommand{\HotHaworWA}{4.70}  %
\newcommand{\HotHaworPA}{8.87}  %
\newcommand{\HotWilorW}{12.48}  %
\newcommand{\HotWilorWA}{3.66}  %
\newcommand{\HotWilorPA}{7.70}  %
\newcommand{\DynW}{5.26}  %
\newcommand{\DynWA}{1.83}  %
\newcommand{\DynPA}{4.15}  %
\newcommand{\DynAddA}{25.8}  %
\newcommand{\DynAddC}{65.4}  %
\newcommand{\DynAddsC}{78.4}  %
\newcommand{\WholeW}{10.41}  %
\newcommand{\WholeWA}{3.26}  %
\newcommand{\WholePA}{6.67}  %
\newcommand{\WholeAddC}{51.1}  %
\newcommand{\WholeAddsC}{69.9}  %
\newcommand{\FpAddA}{12.8}  %
\newcommand{\FpAddC}{38.9}  %
\newcommand{\FpAddsC}{52.8}  %
\newcommand{\FpOursAddA}{11.11}  %
\newcommand{\FpOursAddC}{32.29}  %
\newcommand{\FpOursAddsC}{44.48}  %
\newcommand{\WholeHtoAdd}{44.7}  %
\newcommand{\WholeHtoAdds}{64.9}  %
\newcommand{\HtotrHtoAdd}{62.1}  %
\newcommand{\HtotrHtoAdds}{77.1}  %
\newcommand{\HtoGvPriorAddC}{39.95}  %
\newcommand{\HtoGvPriorAddsC}{64.69}  %
\newcommand{\HtoGvAddC}{64.22}  %
\newcommand{\HtoGvAddsC}{90.97}  %
\newcommand{\HtoSmPriorAddC}{37.35}  %
\newcommand{\HtoSmPriorAddsC}{61.95}  %
\newcommand{\HtoSmAddC}{57.16}  %
\newcommand{\HtoSmAddsC}{83.76}  %
\newcommand{\TestGvPost}{77.62}  %
\newcommand{\TestGvStaged}{76.77}  %
\newcommand{\TestGvJoint}{72.24}  %
\newcommand{\TestGvPostVsJoint}{+5.38 [+4.29, +6.66]}  %
\newcommand{\TestGvPostVsStaged}{+0.85 [+0.48, +1.22]}  %
\newcommand{\TestGvPnpOnly}{69.54}  %
\newcommand{\TestGvPnpRef}{68.98}  %
\newcommand{\TestGvOursVsPnp}{+8.64 [+5.53, +11.52]}  %
\newcommand{\TestSmPost}{66.19}  %
\newcommand{\TestSmStaged}{66.05}  %
\newcommand{\TestSmJoint}{62.75}  %
\newcommand{\TestSmPostVsJoint}{+3.45 [+2.39, +4.48]}  %
\newcommand{\TestSmPnpOnly}{59.94}  %
\newcommand{\TestSmPnpRef}{60.40}  %
\newcommand{\TestSmOursVsPnp}{+5.79 [+3.02, +8.57]}  %
\newcommand{\TestGvPrior}{62.58}  %
\newcommand{\TestSmPrior}{55.36}  %
\newcommand{\TestSmPostVsStagedD}{+0.15}  %
\newcommand{\TestSmPostVsStagedLo}{-0.37}  %
\newcommand{\TestSmPostVsStagedHi}{+0.65}  %
\newcommand{\RobGtGt}{92.8}  %
\newcommand{\RobGtFp}{25.5}  %
\newcommand{\RobGtPrior}{54.9}  %
\newcommand{\RobGtOurs}{58.8}  %
\newcommand{\RobOursGt}{33.3}  %
\newcommand{\RobOursOurs}{30.7}  %
\newcommand{\RobOursVsFpD}{33.3}  %
\newcommand{\RobOursVsFpLo}{19.0}  %
\newcommand{\RobOursVsFpHi}{47.7}  %
\newcommand{\RobPriorVsFpD}{29.4}  %
\newcommand{\RobPriorVsFpLo}{15.0}  %
\newcommand{\RobPriorVsFpHi}{43.1}  %
\newcommand{\ConFOneGv}{92.4}  %
\newcommand{\ConFOneSm}{85.3}  %
\newcommand{\ConOffGvD}{1.77}  %
\newcommand{\ConOffGvLo}{1.30}  %
\newcommand{\ConOffGvHi}{2.30}  %
\newcommand{\ConAlwGvD}{22.55}  %
\newcommand{\ConAlwGvLo}{20.17}  %
\newcommand{\ConAlwGvHi}{25.09}  %
\newcommand{\ConGtGvD}{+0.09}  %
\newcommand{\ConGtGvLo}{-0.01}  %
\newcommand{\ConGtGvHi}{+0.23}  %
\newcommand{\ConOffSmD}{1.27}  %
\newcommand{\ConOffSmLo}{0.84}  %
\newcommand{\ConOffSmHi}{1.78}  %
\newcommand{\ConAlwSmD}{17.86}  %
\newcommand{\ConAlwSmLo}{15.45}  %
\newcommand{\ConAlwSmHi}{20.42}  %
\newcommand{\ConGtSmD}{-0.35}  %
\newcommand{\ConGtSmLo}{-0.85}  %
\newcommand{\ConGtSmHi}{+0.12}  %
\newcommand{\AbFullHotGvP}{64.95}  %
\newcommand{\AbNoPiHotGvP}{60.66}  %
\newcommand{\AbNoDinoHotGvP}{61.13}  %
\newcommand{\AbNoVfmHotGvP}{43.20}  %
\newcommand{\AbNoVfmHotGvPD}{21.8}  %
\newcommand{\AbFullHotGvR}{78.47}  %
\newcommand{\AbNoPiHotGvR}{76.85}  %
\newcommand{\AbNoDinoHotGvR}{75.44}  %
\newcommand{\AbNoVfmHotGvR}{65.53}  %
\newcommand{\AbFullHotSmP}{55.76}  %
\newcommand{\AbNoPiHotSmP}{52.75}  %
\newcommand{\AbNoDinoHotSmP}{52.76}  %
\newcommand{\AbNoVfmHotSmP}{40.75}  %
\newcommand{\AbNoVfmHotSmPD}{15.0}  %
\newcommand{\AbFullHotSmR}{64.58}  %
\newcommand{\AbNoPiHotSmR}{64.14}  %
\newcommand{\AbNoDinoHotSmR}{62.00}  %
\newcommand{\AbNoVfmHotSmR}{55.77}  %
\newcommand{\AbNoDinoHotGvHandD}{0.99}  %
\newcommand{\TestGvPnpFallback}{0.9\%}  %
\newcommand{\TestSmPnpFallback}{16.6\%}  %
\newcommand{\DynSubN}{196}  %
\newcommand{\DynSubGvAddC}{79.21}  %
\newcommand{\DynSubSmAddC}{65.62}  %
\newcommand{\DynSubFpAddC}{32.11}  %
\newcommand{\HotGvAddCSeedB}{77.75}  %
\newcommand{\HotSmAddCSeedB}{64.83}  %
\newcommand{\RtTotA}{26.8}  %
\newcommand{\RtTotC}{50.6}  %
\newcommand{\RtMemModel}{2.9}  %
\newcommand{\SamFailMissPct}{87}  %
\newcommand{\SamGoodDropMax}{0.1}  %
\newcommand{\AbNoPiHotGvRA}{40.74}  %
\newcommand{\AbNoDinoHotGvRA}{38.65}  %
\newcommand{\AbNoVfmHotGvRA}{28.40}  %
\newcommand{\AbFullHotGvRA}{42.73}  %
\newcommand{\AbNoVfmHotGvRAD}{+14.33}  %
\newcommand{\AbNoVfmHotGvRALo}{+11.85}  %
\newcommand{\AbNoVfmHotGvRAHi}{+16.75}  %
\newcommand{\SubGvTruncAdd}{78.21}  %
\newcommand{\SubGvOovAdd}{66.83}  %
\newcommand{\SubGvContactAdd}{78.63}  %
\newcommand{\SubGvPriorTruncAdd}{63.52}  %
\newcommand{\SubGvPriorOovAdd}{49.25}  %
\newcommand{\SubGvPriorContactAdd}{64.85}  %
\newcommand{\WholeTruncAdd}{42.5}  %
\newcommand{\WholeOovAdd}{28.8}  %
\newcommand{\WholeContactAdd}{55.8}  %
\newcommand{\HoiGvAdd}{73.40}  %
\newcommand{\HoiGvPriorAdd}{62.72}  %
\newcommand{\WholeHoiAdd}{53.5}  %
\newcommand{\HotGvAddsA}{76.39}  %
\newcommand{\HotSmAddsA}{56.80}  %
\newcommand{\HotSmPriorAddsA}{39.48}  %
\newcommand{\DynAddsA}{48.2}  %

%% file: sec/X_suppl_arxiv.tex
\clearpage
\maketitlesupplementary
\setcounter{section}{0}\renewcommand{\thesection}{\Alph{section}}\renewcommand{\theHsection}{supp.\Alph{section}}
\setcounter{table}{0}\renewcommand{\thetable}{S\arabic{table}}\renewcommand{\theHtable}{supp.\arabic{table}}
\setcounter{figure}{0}\renewcommand{\thefigure}{S\arabic{figure}}\renewcommand{\theHfigure}{supp.\arabic{figure}}

The supplementary video on the project page (\url{\projecturl}) shows the task and method (0:00--0:30), comparisons on HOT3D (0:30--1:03), zero-shot results on H2O (1:03--1:24), robot retargeting with the reconstructed object (1:24--2:08), more results on HOT3D (2:08--2:28) and failure cases (2:28--2:38).

\section{Implementation details}
\label{sec:supp_impl}

\paragraph{State and tokens.}
The per-frame state $x_t\in\mathbb{R}^{73}$ consists of the object pose (9), the contact of both hands (2) and, for each hand, the global rotation (3), wrist translation (3), finger articulation (15) and shape (10).
The denoiser embeds each frame of the noisy window with a linear layer and adds a sinusoidal temporal encoding, giving 120 tokens per window.
The context encoder receives 36 tokens per frame: 16 resampled $\pi^3$ tokens, 16 resampled DINOv2 tokens, the mask-pooled $\pi^3$ and DINOv2 object tokens, one hand token and one camera token, \ie, 4,320 tokens per window.
Each Perceiver resampler~\cite{alayrac2022flamingo} uses 16 latents, 2 layers and 8 heads.
The template is encoded as a basis point set~\cite{prokudin2019bps} with 1,024 basis points within a radius of 0.30\,m and resampled into 16 template tokens $S$.  %
The camera token encodes the camera center and 6D rotation~\cite{zhou2019continuity} relative to the first camera of the window with Fourier features of 6 frequencies followed by an MLP.
The mask reliability is mapped by an MLP to a per-layer shift and scale of the adaptive layer normalization, initialized to zero so that training starts from the unmodulated denoiser; the reliability is zero for an empty mask.
The prior has 68.4M trainable parameters.  %

\paragraph{Diffusion and sampling.}
We use DDPM~\cite{ho2020ddpm} with 1,000 steps, a linear noise schedule from $10^{-4}$ to $0.02$ and $x_0$ prediction.  %
At inference, we run all 1,000 DDPM sampling steps with the EMA weights and draw a single sample per window without classifier-free guidance.

\paragraph{Training.}
Each training item is a pair of a clip and an object that a hand touches in the clip, and a window of 120 frames is drawn uniformly at random within the clip.
We train from scratch with AdamW at a learning rate of $2\times10^{-4}$ and a weight decay of 0.01, with a linear warmup over the first 1k iterations and a constant learning rate afterwards.  %
We use a batch size of 8, 100k iterations, gradient clipping at a norm of 1.0, bfloat16 mixed precision and an exponential moving average of the weights with a decay of 0.999, and evaluate the final checkpoint.  %
All channels are normalized by the mean and standard deviation of the training split.
Since both backbones are frozen, their features are computed once per clip and cached.
Training the model on 1,311 clips takes about 60 hours on a single NVIDIA RTX 5090 GPU with a peak memory of 14.3\,GB.  %

\paragraph{Long clips.}
For a clip of 150 frames, windows start at frames 0 and 30.
Each window is sampled and refined independently, and the two windows are then merged in the world frame.
In the overlap, the weight of the later window rises from 0 to 1 along a raised-cosine ramp; rotations are blended by spherical interpolation, and translations and finger articulations linearly, while the contact of each frame is taken from the window with the larger weight.
The merged poses are smoothed by a temporal Gaussian filter with $\sigma=2$ frames (radius 6, reflective padding), and the hand shape is taken from the first frame of the first window.
The analyses on 1,000 windows (\cref{tab:supp_analysis}) evaluate single windows of 120 frames and therefore involve no merging.

\section{Loss and refinement hyperparameters}
\label{sec:supp_hparams}

\begin{table*}[t]
\centering
\caption{\textbf{Loss and refinement hyperparameters.} The same values are used for all models and on both datasets.}
\label{tab:supp_hparams}
\footnotesize
\setlength{\tabcolsep}{4pt}
\begin{tabular}{ll}
\toprule
\multicolumn{2}{l}{\emph{Training loss}} \\
Channel weights $w_j$ & object 5, contact 5, hand 1, shape 0.1 \\  %
$\lambda_\mathrm{rel}$ & $4.1\times10^{-3}$, from 10k iter., linear ramp over 5k \\  %
Mask dropout & window $p=0.15$; segments $p=0.35$, 1--60 frames \\  %
Template dropout & $p=0.15$ \\  %
\midrule
\multicolumn{2}{l}{\emph{Refinement}} \\
Phase 1 (translation) & 200 Adam steps, lr 0.002 \\  %
Phase 2 (rotation, translation) & 200 Adam steps, lr 0.005 / 0.002 \\  %
$\lambda_\mathrm{sil}$, $\lambda_\mathrm{prior}$, $\lambda_\mathrm{smooth}$, $\lambda_\mathrm{static}$ & 1, 0.01, 0.1, 1 \\  %
Rotation smoothness, prior & 1, 0.1 \\  %
Template points, mask samples & 2,048, 1,024 \\
Contact threshold, min.\ segment & 0.5, 2 frames \\  %
\bottomrule
\end{tabular}
\end{table*}

\paragraph{Training loss.}
\cref{tab:supp_hparams} lists the weights of the training loss.
The channel weights $w_j$ of \cref{eq:diff} are shared within each group of channels, \ie, the 9 object channels, the 2 contact channels, the 42 global-pose, translation and articulation channels of the two hands, and the 20 shape channels.
The weight $\lambda_\mathrm{rel}$ is zero for the first 10k iterations and then increases linearly to its final value over 5k iterations.
In $\mathcal{L}_\mathrm{rel}$, translations are measured in meters.
Mask segments are dropped independently of the whole-window dropout, with a length drawn log-uniformly, and a dropped template is replaced by a learned null template.

\paragraph{Refinement.}
Distances in the energy are measured in centimeters.
The silhouette term uses 2,048 points sampled on the template surface and 1,024 points sampled on the mask boundary, with the masks resized to $352\times352$ on HOT3D and $320\times180$ on H2O.
The first phase optimizes the per-frame translation with the rotation fixed to the sample.
We optimize the translation first because the mask energy constrains the translation of the sample most directly.
The second phase starts from the result of the first and optimizes the translation together with a rotation update, parameterized as a right-multiplied increment in $\mathfrak{so}(3)$ initialized to zero, with separate learning rates for rotation and translation.
Both phases use the same translation terms, and the second phase adds the rotation smoothness and the rotation prior.
The static term acts only on the object center.
The energy weights were selected on the development windows (\cref{sec:supp_splits}) and kept fixed for all experiments, including H2O.

\section{Inputs and preprocessing}
\label{sec:supp_inputs}

\paragraph{Images and backbones.}
The fisheye images of HOT3D are rectified to a pinhole model with the calibration of each recording, rotated upright and resized to $224\times224$, giving $16\times16$ patches of 14 pixels.
H2O frames are resized to the same resolution.
For $\pi^3$~\cite{wang2025pi3}, we use the released model and concatenate the decoder features of its last two blocks into a 2,048-dimensional feature per patch; the point and camera heads are not used, and attention across frames is computed within chunks of 16 frames.
For DINOv2~\cite{oquab2024dinov2}, we use the ViT-L/14 model with registers and take its normalized 1,024-dimensional patch tokens.
Both backbones are frozen.

\paragraph{Mask pooling.}
The object mask is reduced to the patch grid as a soft coverage $w_p\in[0,1]$, the fraction of each $14\times14$ patch covered by the mask.
The object token of each backbone is the coverage-weighted mean $\sum_p w_p f_p/\sum_p w_p$ of its patch features $f_p$, followed by a linear projection.

\paragraph{Hand track.}
We run HaWoR~\cite{zhang2025hawor} on the rectified frames.
Instead of estimating the camera with its SLAM module, we lift its camera-frame MANO estimates to the world frame with the same camera poses that are given to all methods.
The hand track uses the mean MANO shape.
Its finger articulation is then converted into the 15 PCA coefficients of our state.
The hand tracks of the training clips are obtained in the same way, so that the prior receives the same converted track in training and evaluation.
The HaWoR rows of \cref{tab:hot3d,tab:supp_hot3d} report the native estimates of HaWoR instead.

\paragraph{Ground-truth visible masks.}
On HOT3D, we use the visible (modal) object masks provided with the dataset.

\paragraph{Predicted masks.}
For \samthree{}~\cite{carion2026sam3}, the text prompt is the name of the object.
The names were written once from renderings of the templates alone, before any result was inspected, and were kept fixed afterwards.
On HOT3D, \samthree{} tracks the object through all 150 frames of a clip at a resolution of 448 pixels, the output probability is thresholded at 0.5, and the track with the highest summed probability is used.  %
The same resolution, threshold and prompts are used on H2O.

\section{Data splits}
\label{sec:supp_splits}

\paragraph{HOT3D.}
The Aria clips of HOT3D with public ground truth comprise 1,516 clips of 150 frames from 124 recordings.
Of these, 25 recordings are not part of the training data of HaWoR~\cite{zhang2025hawor}, and all our evaluation and development data come from them.
Our 147 evaluation clips cover 10 of these recordings entirely.
The recordings are taken in a fixed random order so that the number of clips matches the test set of DynamicHOI~\cite{guo2026dynamichoi}.
Five further recordings (49 clips) form the development set and are excluded from training.
To match the training-set size of DynamicHOI, the model of the main results is trained on 1,311 clips sampled at random with seed 0 from the remaining recordings.
On the 147 clips, we evaluate all 228 objects touched by a hand, identified from the ground-truth contact.
\cref{tab:supp_recordings} lists the evaluation and development recordings.

\begin{table*}[t]
\centering
\caption{\textbf{HOT3D recordings} of our evaluation and development sets.}
\label{tab:supp_recordings}
\footnotesize
\begin{tabular}{ll}
\toprule
Evaluation (147 clips) & P0001\_a68492d5, P0002\_2ea9af5b, P0010\_0ecbf39f, \\
 & P0010\_1c9fe708, P0011\_11475e24, P0011\_76ea6d47, \\
 & P0012\_130a66e1, P0012\_d85e10f6, P0012\_f7e3880b, \\
 & P0014\_8254f925 \\
\midrule
Development & P0001\_9b6feab7, P0010\_41c4c626, P0011\_cee8fe4f, \\
 & P0012\_476bae57, P0015\_42b8b389 \\
\bottomrule
\end{tabular}
\end{table*}

\paragraph{Analysis split.}
The analyses of \cref{tab:supp_analysis} use a model trained on the 1,167 clips of the other 99 recordings, which keeps all 25 recordings held out.
The 25 held-out recordings are divided with a fixed seed into the 5 development recordings above and 20 test recordings, which include the 10 evaluation recordings.
Taking windows of 120 frames starting at frames 0 and 30 for every object touched by a hand in each clip gives 1,000 test windows and 178 development windows.
The 1,311-clip model is trained on clips of some of these 20 recordings but on none of the 10 evaluation recordings, so the analyses on these windows use the 1,167-clip model.

\paragraph{H2O.}
We use the 218 clips of the official test split of H2O~\cite{kwon2021h2o}.
One clip has no ground-truth object pose and is excluded, leaving 217 clips, each evaluated with two windows merged as on HOT3D.

\section{Evaluation protocol}
\label{sec:supp_eval}

\paragraph{Hand metrics.}
For W and WA, we fit a similarity transform to all joints of the first two valid frames and of all valid frames, respectively, and report the MPJPE over all valid frames in centimeters.
PA aligns each frame separately by Procrustes analysis and is reported in millimeters.
Each hand is scored once per evaluated object track, and hands with fewer than 5 valid frames are skipped; the scores are then averaged.
Ground-truth joints use the ground-truth MANO shape, and the predicted shape is that of the first frame of the first window.

\paragraph{Object metrics.}
ADD is the mean distance between corresponding template vertices under the predicted and ground-truth poses, and ADD-S the mean distance from each predicted vertex to the nearest vertex of the ground-truth mesh.
We report the area under the curve of the fraction of frames below a distance threshold, for thresholds up to 0.1\,m and 0.3\,m, pooling the valid frames of all object tracks; FoundationPose and our method are scored on the same frames.  %
The area is divided by the maximum threshold and reported in percent.
For \cref{tab:subsets}, contact means ground-truth hand--object contact, and truncated and out of view mean that some but not all, or none, of the projected ground-truth template lies inside the image. For the HOI column, following WHOLE~\cite{ye2026whole}, we fit one similarity transform to the joints of both hands, apply it to the predicted object and then compute ADD.

\paragraph{Baselines on our clips.}
FoundationPose~\cite{wen2024foundationpose} is given the same ground-truth visible masks, templates and camera poses as ours, with metric depth from Depth Anything~3~\cite{lin2025depthanything3}.
It is registered once on the first frame whose mask covers at least 100 pixels and tracked afterwards. Frames without a FoundationPose estimate, including those during occlusion, count as failures in its object scores.
HaWoR and WiLoR are lifted to the world frame with the same camera poses and scored with the same protocol on the same clips.

\paragraph{Confidence intervals.}
For the analyses on 1,000 windows, 95\% confidence intervals come from a recording-level cluster bootstrap with 2,000 resamples, and we report the widest of the percentile, BCa and jackknife intervals.  %
For the feature ablations and the contact analysis, we use a paired clip-level bootstrap with 10,000 resamples, and for the robot experiments a paired track-level bootstrap with 10,000 resamples over the success rate averaged across seeds.

\section{Full results on HOT3D}
\label{sec:supp_full}
\cref{tab:supp_hot3d} reports all hand and object metrics on HOT3D for both mask settings.
The selection rule of WHOLE, an object displacement above 5\,cm from the first frame, keeps \DynSubN{} of our 228 tracks; on them, the ADD@.3 of \method{} is \DynSubGvAddC{} with ground-truth visible masks and \DynSubSmAddC{} with \samthree{} masks, and that of FoundationPose$^\dagger$ is \DynSubFpAddC{}.  %
A second training seed of the main model reaches an ADD@.3 of \HotGvAddCSeedB{} with ground-truth visible masks and \HotSmAddCSeedB{} with \samthree{} masks, compared with \HotGvAddC{} and \HotSmAddC{} for the reported seed.

\begin{table*}[t]
\centering
\caption{\textbf{Full hand and object results on HOT3D} for both mask settings. $^*$Paper-reported results; the evaluation clips differ from ours (FoundationPose as reproduced by DynamicHOI). $^\dagger$Evaluated on our 147 clips; HaWoR is its native output, lifted to the world with the given camera poses, as in \cref{tab:hot3d}. Ours shaded.}  %
\label{tab:supp_hot3d}
\small
\setlength{\tabcolsep}{4pt}
\begin{tabular}{llcccccc}
\toprule
 & & \multicolumn{3}{c}{Hand} & \multicolumn{3}{c}{Object} \\
\cmidrule(lr){3-5}\cmidrule(lr){6-8}
Method & Mask & W$\downarrow$ & WA$\downarrow$ & PA$\downarrow$ & ADD@.1$\uparrow$ & ADD@.3$\uparrow$ & ADD-S@.3$\uparrow$ \\
\midrule
HaWoR$^\dagger$~\cite{zhang2025hawor} & -- & \HotHaworW & \HotHaworWA & \HotHaworPA & -- & -- & -- \\
WiLoR$^\dagger$~\cite{potamias2025wilor} & -- & \HotWilorW & \HotWilorWA & \HotWilorPA & -- & -- & -- \\
FoundationPose$^\dagger$~\cite{wen2024foundationpose} & GT visible & -- & -- & -- & \FpOursAddA & \FpOursAddC & \FpOursAddsC \\
FoundationPose$^*$~\cite{wen2024foundationpose} & -- & -- & -- & -- & \FpAddA & \FpAddC & \FpAddsC \\
WHOLE$^*$~\cite{ye2026whole} & -- & \WholeW & \WholeWA & \WholePA & -- & \WholeAddC & \WholeAddsC \\
DynamicHOI$^*$~\cite{guo2026dynamichoi} & -- & \DynW & \DynWA & \DynPA & \DynAddA & \DynAddC & \DynAddsC \\
\midrule
\oursrow \method{} (prior) & \samthree{} & \HotSmW & \HotSmWA & \HotSmPA & \HotSmPriorAddA & \HotSmPriorAddC & \HotSmPriorAddsC \\
\oursrow \method{} & \samthree{} & \HotSmW & \HotSmWA & \HotSmPA & \HotSmAddA & \HotSmAddC & \HotSmAddsC \\
\oursrow \method{} (prior) & GT visible & \HotGvW & \HotGvWA & \HotGvPA & \HotGvPriorAddA & \HotGvPriorAddC & \HotGvPriorAddsC \\
\oursrow \method{} & GT visible & \HotGvW & \HotGvWA & \HotGvPA & \HotGvAddA & \HotGvAddC & \HotGvAddsC \\
\bottomrule
\end{tabular}
\end{table*}

\section{Initialization and guidance analysis}
\label{sec:supp_analysis}

\begin{table}[t]
\centering
\caption{\textbf{Initialization and observation fitting.} This analysis uses a separate model trained on 1,167 clips and 1,000 held-out HOT3D windows of 120 frames; its values are not comparable to the main results (1,311-clip model, 147 clips of 150 frames). Guidance applies the same data terms inside the sampling loop, updating rotation and translation jointly or following our schedule (translation, then full pose). In the PnP rows, only the object trajectory is replaced: the hands and contact are those sampled by the prior, and windows in which PnP fails (\TestGvPnpFallback{} with ground-truth visible masks, \TestSmPnpFallback{} with \samthree{} masks) keep the object trajectory of the prior. ``Default'' is the configuration of \method{}. ADD@.3$\uparrow$; best in bold.}
\label{tab:supp_analysis}
\footnotesize
\setlength{\tabcolsep}{4pt}
\begin{tabular}{llcc}
\toprule
Init. & Observation fitting & GT vis. & \samthree{} \\
\midrule
PnP & -- & \TestGvPnpOnly & \TestSmPnpOnly \\
PnP & refinement & \TestGvPnpRef & \TestSmPnpRef \\
\midrule
Prior & -- & \TestGvPrior & \TestSmPrior \\
Prior & joint guidance & \TestGvJoint & \TestSmJoint \\
Prior & staged guidance & \TestGvStaged & \TestSmStaged \\
Prior & refinement (default) & \textbf{\TestGvPost} & \textbf{\TestSmPost} \\
\bottomrule
\end{tabular}
\end{table}

\paragraph{Prior versus PnP initialization.}
To assess the role of object initialization, we replace only the sampled object trajectory with a training-free perspective-n-point (PnP) estimate in the spirit of FoundPose~\cite{ornek2024foundpose}.
It matches DINOv2 features of the masked object crop to renderings of the textured template, and we then apply the same refinement with the hands and contact sampled by the prior.
As shown in \cref{tab:supp_analysis}, the refinement barely changes the PnP initialization, whereas the same refinement raises the prior well above it.
As a result, \method{} outperforms the PnP-initialized variant by \TestGvOursVsPnp{} ADD@.3 with ground-truth visible masks and \TestSmOursVsPnp{} with \samthree{} masks.
Note that PnP additionally requires a textured template.

\paragraph{Guidance versus post-sampling refinement.}
With our prior, a single refinement after sampling outperforms guidance that updates rotation and translation jointly inside the loop by \TestGvPostVsJoint{} and \TestSmPostVsJoint{} ADD@.3, and matches or slightly exceeds guidance that follows the same schedule (\TestGvPostVsStaged{} and $\TestSmPostVsStagedD$ [$\TestSmPostVsStagedLo$, $\TestSmPostVsStagedHi$]).
The joint and staged guidance schedules use 1,000 and 5,000 gradient steps, respectively, versus 400 for post-sampling refinement; both schedules were selected on the development windows.  %
We use the refinement after sampling, since it leaves the sampler unchanged and adds only a short optimization after it.

\section{Results with predicted masks}
\label{sec:supp_sam}
This section repeats the analyses of the main paper with masks predicted by \samthree{}~\cite{carion2026sam3} from a text prompt that names the object, alongside the ground-truth visible masks. In each setting, the same masks are used as model input and in refinement.

\begin{table}[t]
\centering
\caption{\textbf{Effect of the frozen foundation features} under both mask settings (ADD@.3$\uparrow$, prior / after refinement; best in bold). All models use the same 1,311 clips, training seed and schedule; each ablation omits the corresponding feature tokens and extends \cref{tab:ablation}.}
\label{tab:supp_ablation}
\footnotesize
\setlength{\tabcolsep}{4pt}
\begin{tabular}{lcccc}
\toprule
 & \multicolumn{4}{c}{HOT3D} \\
\cmidrule(lr){2-5}
 & \multicolumn{2}{c}{GT vis.} & \multicolumn{2}{c}{\samthree{}} \\
Model & prior & ref. & prior & ref. \\
\midrule
\method{} & \textbf{\AbFullHotGvP} & \textbf{\AbFullHotGvR} & \textbf{\AbFullHotSmP} & \textbf{\AbFullHotSmR} \\
w/o $\pi^3$ & \AbNoPiHotGvP & \AbNoPiHotGvR & \AbNoPiHotSmP & \AbNoPiHotSmR \\
w/o DINOv2 & \AbNoDinoHotGvP & \AbNoDinoHotGvR & \AbNoDinoHotSmP & \AbNoDinoHotSmR \\
w/o $\pi^3$, DINOv2 & \AbNoVfmHotGvP & \AbNoVfmHotGvR & \AbNoVfmHotSmP & \AbNoVfmHotSmR \\
\bottomrule
\end{tabular}
\end{table}

\paragraph{Foundation features.}
\cref{tab:supp_ablation} repeats the feature ablation of \cref{tab:ablation} with \samthree{} masks.
With \samthree{} masks, removing both feature sets lowers the ADD@.3 of the prior on HOT3D by \AbNoVfmHotSmPD{} points.

\paragraph{Contact in the refinement.}
On the 147 HOT3D clips, the contact sampled by the prior agrees with the ground truth at an F1 score of \ConFOneGv{}\% with ground-truth visible masks and \ConFOneSm{}\% with \samthree{} masks.  %
Replacing it with ground-truth contact changes ADD@.3 by only $\ConGtGvD$ [$\ConGtGvLo$, $\ConGtGvHi$] and $\ConGtSmD$ [$\ConGtSmLo$, $\ConGtSmHi$] percentage points with ground-truth visible and \samthree{} masks, respectively.
Removing the static term lowers ADD@.3 by \ConOffGvD{} [\ConOffGvLo{}, \ConOffGvHi{}] and \ConOffSmD{} [\ConOffSmLo{}, \ConOffSmHi{}], and ignoring contact by treating every object as resting lowers it by \ConAlwGvD{} [\ConAlwGvLo{}, \ConAlwGvHi{}] and \ConAlwSmD{} [\ConAlwSmLo{}, \ConAlwSmHi{}], respectively.

\begin{figure}[t]
\centering
\IfFileExists{fig/fig7_sam_masks.jpg}{\includegraphics[width=\linewidth]{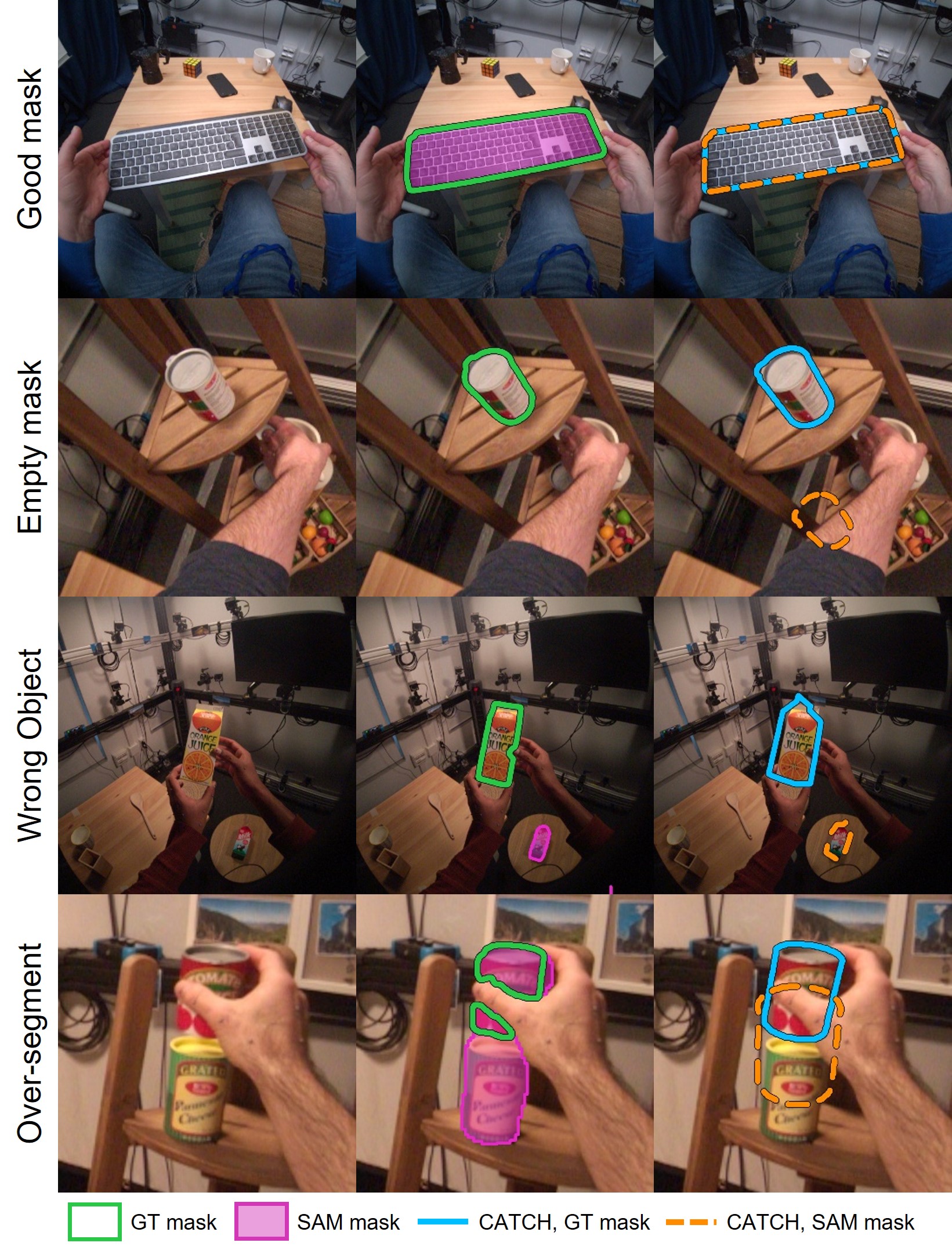}}{\fbox{Fig.~7 placeholder}}
\caption{\textbf{When predicted masks fail} on HOT3D. Representative tracks, from top: a good mask, an empty mask, a mask on the wrong object and an over-segmented mask. From left: the input frame; the ground-truth visible mask (green outline) and the \samthree{} mask (magenta); the final object of \method{} projected with ground-truth visible masks (blue) and with \samthree{} masks (orange, dashed).}
\label{fig:supp_sam_failure}
\end{figure}

\paragraph{When predicted masks fail.}
Comparing each HOT3D object track under the two mask settings, the large drops in ADD@.3 come mostly from tracks in which \samthree{} misses the object or segments a different one: they make up \SamFailMissPct{}\% of the tracks that lose at least 5 points, most of them with an empty mask.  %
Where the \samthree{} mask agrees with the ground truth (mean IoU of at least 0.8), the average drop is \SamGoodDropMax{} points.  %
\cref{fig:supp_sam_failure} shows a representative track for each case.

\section{Robot retargeting details}
\label{sec:supp_robot}
We retarget the trajectories to one or two XHAND robot hands with SPIDER~\cite{pan2026spider} in physics simulation.
Of the 228 evaluated tracks, 161 pass task eligibility: the object moves more than 10\,cm, a hand touches it for at least 30 frames, and it does not start below the simulation table.
We retain 51 tracks whose ground-truth retargeting succeeds in at least two of the seeds 0, 1 and 2 and whose unmoved object does not already satisfy the goal.
Selection uses only ground truth, and every row of \cref{tab:robot} is run on the same tracks with the same three seeds.
A rollout succeeds when the final object position lies within 10\,cm of the ground truth.
With ground-truth hands, the objects of \method{} succeed \RobOursVsFpD{} [\RobOursVsFpLo{}, \RobOursVsFpHi{}] points more often than those of FoundationPose, and samples of the prior \RobPriorVsFpD{} [\RobPriorVsFpLo{}, \RobPriorVsFpHi{}] points more often.
In all rows, the simulation starts from the ground-truth first frame.
The wrist orientation of every input hand is computed from its own joints; only the constant rotation that maps this joint-based frame to the MANO wrist frame is fitted per clip and hand on the ground-truth hand, independently of the reconstruction.
FoundationPose is run as in \cref{sec:supp_eval}; for retargeting, its gaps are interpolated in the world frame and the first and last estimated poses are held. The hands of \method{} are the same as in \cref{tab:hot3d}.

\section{Runtime}
\label{sec:supp_runtime}
On a single NVIDIA RTX 5090 GPU, sampling and refinement take \RtTotA{}\,s for a HOT3D clip of 150 frames with one object and \RtTotC{}\,s with three objects (median of three runs, model loading excluded), with a peak GPU memory of \RtMemModel{}\,GB.  %
The two frozen backbones take less than a second per clip.

%% file: arxiv.bbl
\begin{thebibliography}{68}
\providecommand{\natexlab}[1]{#1}
\providecommand{\url}[1]{\texttt{#1}}
\expandafter\ifx\csname urlstyle\endcsname\relax
  \providecommand{\doi}[1]{doi: #1}\else
  \providecommand{\doi}{doi: \begingroup \urlstyle{rm}\Url}\fi

\bibitem[Alayrac et~al.(2022)Alayrac, Donahue, Luc, Miech, Barr, Hasson, Lenc, Mensch, Millican, Reynolds, Ring, Rutherford, Cabi, Han, Gong, Samangooei, Monteiro, Menick, Borgeaud, Brock, Nematzadeh, Sharifzadeh, Bi\'{n}kowski, Barreira, Vinyals, Zisserman, and Simonyan]{alayrac2022flamingo}
Jean-Baptiste Alayrac, Jeff Donahue, Pauline Luc, Antoine Miech, Iain Barr, Yana Hasson, Karel Lenc, Arthur Mensch, Katherine Millican, Malcolm Reynolds, Roman Ring, Eliza Rutherford, Serkan Cabi, Tengda Han, Zhitao Gong, Sina Samangooei, Marianne Monteiro, Jacob~L Menick, Sebastian Borgeaud, Andy Brock, Aida Nematzadeh, Sahand Sharifzadeh, Miko\l~aj Bi\'{n}kowski, Ricardo Barreira, Oriol Vinyals, Andrew Zisserman, and Kar\'{e}n Simonyan.
\newblock Flamingo: a visual language model for few-shot learning.
\newblock In \emph{Advances in Neural Information Processing Systems}, pages 23716--23736. Curran Associates, Inc., 2022.

\bibitem[Aytekin et~al.(2026)Aytekin, Chen, Shen, Beeler, Rhodin, Dabral, and Theobalt]{aytekin2026grag}
Ayce~Idil Aytekin, Xu Chen, Zhengyang Shen, Thabo Beeler, Helge Rhodin, Rishabh Dabral, and Christian Theobalt.
\newblock Grasp in {Gaussians}: Fast monocular reconstruction of dynamic hand-object interactions.
\newblock \emph{arXiv preprint arXiv:2604.12929}, 2026.

\bibitem[Banerjee et~al.(2025)Banerjee, Shkodrani, Moulon, Hampali, Han, Zhang, Zhang, Fountain, Miller, Basol, Newcombe, Wang, Engel, and Hodan]{banerjee2025hot3d}
Prithviraj Banerjee, Sindi Shkodrani, Pierre Moulon, Shreyas Hampali, Shangchen Han, Fan Zhang, Linguang Zhang, Jade Fountain, Edward Miller, Selen Basol, Richard~A. Newcombe, Robert Wang, Jakob~Julian Engel, and Tomas Hodan.
\newblock {HOT3D:} hand and object tracking in {3D} from egocentric multi-view videos.
\newblock In \emph{{IEEE/CVF} Conference on Computer Vision and Pattern Recognition, {CVPR} 2025, Nashville, TN, USA, June 11-15, 2025}, pages 7061--7071. Computer Vision Foundation / {IEEE}, 2025.

\bibitem[Bansal et~al.(2026)Bansal, Zhu, Tripathi, Zhao, Black, and Damen]{bansal2026hopformer}
Siddhant Bansal, Zhifan Zhu, Shashank Tripathi, Jiahe Zhao, Michael~J. Black, and Dima Damen.
\newblock Towards in-the-wild egocentric {3D} hand-object pose estimation.
\newblock In \emph{European Conference on Computer Vision (ECCV)}, pages 21--39, 2026.

\bibitem[Cao et~al.(2021)Cao, Radosavovic, Kanazawa, and Malik]{cao2021mow}
Zhe Cao, Ilija Radosavovic, Angjoo Kanazawa, and Jitendra Malik.
\newblock Reconstructing hand-object interactions in the wild.
\newblock In \emph{2021 {IEEE/CVF} International Conference on Computer Vision, {ICCV} 2021, Montreal, QC, Canada, October 10-17, 2021}, pages 12397--12406. {IEEE}, 2021.

\bibitem[Carion et~al.(2026)Carion, Gustafson, Hu, Debnath, Hu, Suris, Ryali, Alwala, Khedr, Huang, Lei, Ma, Guo, Kalla, Marks, Greer, Wang, Sun, Rädle, Afouras, Mavroudi, Xu, Wu, Zhou, Momeni, Hazra, Ding, Vaze, Porcher, Li, Li, Kamath, Cheng, Dollár, Ravi, Saenko, Zhang, and Feichtenhofer]{carion2026sam3}
Nicolas Carion, Laura Gustafson, Yuan-Ting Hu, Shoubhik Debnath, Ronghang Hu, Didac Suris, Chaitanya Ryali, Kalyan~Vasudev Alwala, Haitham Khedr, Andrew Huang, Jie Lei, Tengyu Ma, Baishan Guo, Arpit Kalla, Markus Marks, Joseph Greer, Meng Wang, Peize Sun, Roman Rädle, Triantafyllos Afouras, Effrosyni Mavroudi, Katherine Xu, Tsung-Han Wu, Yu Zhou, Liliane Momeni, Rishi Hazra, Shuangrui Ding, Sagar Vaze, Francois Porcher, Feng Li, Siyuan Li, Aishwarya Kamath, Ho~Kei Cheng, Piotr Dollár, Nikhila Ravi, Kate Saenko, Pengchuan Zhang, and Christoph Feichtenhofer.
\newblock {SAM} 3: Segment anything with concepts.
\newblock In \emph{International Conference on Learning Representations (ICLR)}, 2026.

\bibitem[Chen et~al.(2026{\natexlab{a}})Chen, Dong, Wu, Wang, Jangir, Niu, Ye, Bharadhwaj, Erickson, and Ichnowski]{chen2026manipulation}
Hongyi Chen, Tony Dong, Tiancheng Wu, Liquan Wang, Yash Jangir, Yaru Niu, Yufei Ye, Homanga Bharadhwaj, Zackory Erickson, and Jeffrey Ichnowski.
\newblock Dexterous manipulation policies from {RGB} human videos via {3D} hand-object trajectory reconstruction.
\newblock \emph{arXiv preprint arXiv:2602.09013}, 2026{\natexlab{a}}.

\bibitem[Chen et~al.(2026{\natexlab{b}})Chen, Shao, Ji, Yang, and Mu]{chen2026v2pmanip}
Kaihan Chen, Yanming Shao, Haifeng Ji, Xiaokang Yang, and Yao Mu.
\newblock {V2P-Manip}: Learning dexterous manipulation from monocular human videos.
\newblock \emph{arXiv preprint arXiv:2606.16436}, 2026{\natexlab{b}}.

\bibitem[Chen et~al.(2026{\natexlab{c}})Chen, Chang, Ye, Zhang, Fang, Li, and Han]{chen2026forehoi}
Yuantao Chen, Jiahao Chang, Chongjie Ye, Chaoran Zhang, Zhaojie Fang, Chenghong Li, and Xiaoguang Han.
\newblock {ForeHOI}: Feed-forward {3D} object reconstruction from daily hand-object interaction videos.
\newblock In \emph{Proceedings of the IEEE/CVF Conference on Computer Vision and Pattern Recognition (CVPR)}, pages 8868--8879, 2026{\natexlab{c}}.

\bibitem[Chen et~al.(2022)Chen, Hasson, Schmid, and Laptev]{chen2022alignsdf}
Zerui Chen, Yana Hasson, Cordelia Schmid, and Ivan Laptev.
\newblock {AlignSDF}: Pose-aligned signed distance fields for hand-object reconstruction.
\newblock In \emph{Computer Vision - {ECCV} 2022 - 17th European Conference, Tel Aviv, Israel, October 23-27, 2022, Proceedings, Part {I}}, pages 231--248. Springer, 2022.

\bibitem[Chen et~al.(2025)Chen, Potamias, Chen, and Schmid]{chen2025hort}
Zerui Chen, Rolandos~Alexandros Potamias, Shizhe Chen, and Cordelia Schmid.
\newblock {HORT}: Monocular hand-held objects reconstruction with transformers.
\newblock In \emph{Proceedings of the IEEE/CVF International Conference on Computer Vision (ICCV)}, pages 6046--6057, 2025.

\bibitem[Cho et~al.(2023)Cho, Kim, Kim, Lee, Ismayilzada, and Baek]{cho2023transformer}
Hoseong Cho, Chanwoo Kim, Jihyeon Kim, Seongyeong Lee, Elkhan Ismayilzada, and Seungryul Baek.
\newblock Transformer-based unified recognition of two hands manipulating objects.
\newblock In \emph{Proceedings of the IEEE/CVF Conference on Computer Vision and Pattern Recognition (CVPR)}, pages 4769--4778, 2023.

\bibitem[Ehsani et~al.(2020)Ehsani, Tulsiani, Gupta, Farhadi, and Gupta]{ehsani2020force}
Kiana Ehsani, Shubham Tulsiani, Saurabh Gupta, Ali Farhadi, and Abhinav Gupta.
\newblock {Use the Force, Luke! Learning} to predict physical forces by simulating effects.
\newblock In \emph{Proceedings of the IEEE/CVF Conference on Computer Vision and Pattern Recognition (CVPR)}, 2020.

\bibitem[Fan et~al.(2024)Fan, Parelli, Kadoglou, Chen, Kocabas, Black, and Hilliges]{fan2024hold}
Zicong Fan, Maria Parelli, Maria~Eleni Kadoglou, Xu Chen, Muhammed Kocabas, Michael~J. Black, and Otmar Hilliges.
\newblock {HOLD:} category-agnostic {3D} reconstruction of interacting hands and objects from video.
\newblock In \emph{{IEEE/CVF} Conference on Computer Vision and Pattern Recognition, {CVPR} 2024, Seattle, WA, USA, June 16-22, 2024}, pages 494--504. {IEEE}, 2024.

\bibitem[Fu et~al.(2026)Fu, Wang, Qiao, Potamias, Komura, Yang, Liu, and Zhao]{fu2026egograsp}
Hongming Fu, Wenjia Wang, Xiaozhen Qiao, Rolandos~Alexandros Potamias, Taku Komura, Shuo Yang, Zheng Liu, and Bo Zhao.
\newblock {EgoGrasp}: World-space hand-object interaction estimation from egocentric videos.
\newblock \emph{arXiv preprint arXiv:2601.01050}, 2026.

\bibitem[Grady et~al.(2021)Grady, Tang, Twigg, Vo, Brahmbhatt, and Kemp]{grady2021contactopt}
Patrick Grady, Chengcheng Tang, Christopher~D. Twigg, Minh Vo, Samarth Brahmbhatt, and Charles~C. Kemp.
\newblock {ContactOpt}: Optimizing contact to improve grasps.
\newblock In \emph{{IEEE} Conference on Computer Vision and Pattern Recognition, {CVPR} 2021, virtual, June 19-25, 2021}, pages 1471--1481. Computer Vision Foundation / {IEEE}, 2021.

\bibitem[Guo et~al.(2026)Guo, Huang, and Kong]{guo2026dynamichoi}
Wenliang Guo, Zhanbo Huang, and Yu Kong.
\newblock {DynamicHOI}: Coupled dynamics for physics-aware {HOI} reconstruction.
\newblock \emph{arXiv preprint arXiv:2609.36454}, 2026.

\bibitem[Hasson et~al.(2019)Hasson, Varol, Tzionas, Kalevatykh, Black, Laptev, and Schmid]{hasson2019obman}
Yana Hasson, G{\"{u}}l Varol, Dimitrios Tzionas, Igor Kalevatykh, Michael~J. Black, Ivan Laptev, and Cordelia Schmid.
\newblock Learning joint reconstruction of hands and manipulated objects.
\newblock In \emph{{IEEE} Conference on Computer Vision and Pattern Recognition, {CVPR} 2019, Long Beach, CA, USA, June 16-20, 2019}, pages 11807--11816. Computer Vision Foundation / {IEEE}, 2019.

\bibitem[Ho et~al.(2020)Ho, Jain, and Abbeel]{ho2020ddpm}
Jonathan Ho, Ajay Jain, and Pieter Abbeel.
\newblock Denoising diffusion probabilistic models.
\newblock In \emph{Advances in Neural Information Processing Systems 33: Annual Conference on Neural Information Processing Systems 2020, NeurIPS 2020, December 6-12, 2020, virtual}, 2020.

\bibitem[Hu et~al.(2022)Hu, Yi, Zhang, Yong, and Xu]{hu2022physical}
Haoyu Hu, Xinyu Yi, Hao Zhang, Jun{-}Hai Yong, and Feng Xu.
\newblock Physical interaction: Reconstructing hand-object interactions with physics.
\newblock In \emph{{SIGGRAPH} Asia 2022 Conference Papers, {SA} 2022, Daegu, Republic of Korea, December 6-9, 2022}, pages 43:1--43:9. {ACM}, 2022.

\bibitem[Ji et~al.(2026)Ji, Pang, Li, Hu, Wang, Sun, Cao, Dai, Luu, Ma, Chen, Cao, Hong, Chen, and Liu]{ji2026egoxperience}
Xinhao Ji, Hui~En Pang, Hao Li, Kairui Hu, Jiner Wang, Mingze Sun, Ziang Cao, Yalun Dai, Thang Luu, Qin Ma, Hongyi Chen, Yukang Cao, Fangzhou Hong, Zhaoxi Chen, and Ziwei Liu.
\newblock {EgoXperience}: Learning world-grounded egocentric representations from human experience.
\newblock \emph{arXiv preprint arXiv:2610.26625}, 2026.

\bibitem[Karunratanakul et~al.(2020)Karunratanakul, Yang, Zhang, Black, Muandet, and Tang]{karunratanakul2020grasping}
Korrawe Karunratanakul, Jinlong Yang, Yan Zhang, Michael~J. Black, Krikamol Muandet, and Siyu Tang.
\newblock Grasping field: Learning implicit representations for human grasps.
\newblock In \emph{2020 International Conference on 3D Vision (3DV)}, pages 333--344. IEEE, 2020.

\bibitem[Kwon et~al.(2021)Kwon, Tekin, St\"uhmer, Bogo, and Pollefeys]{kwon2021h2o}
Taein Kwon, Bugra Tekin, Jan St\"uhmer, Federica Bogo, and Marc Pollefeys.
\newblock {H2O}: Two hands manipulating objects for first person interaction recognition.
\newblock In \emph{Proceedings of the IEEE/CVF International Conference on Computer Vision (ICCV)}, pages 10138--10148, 2021.

\bibitem[Labb{\'{e}} et~al.(2023)Labb{\'{e}}, Manuelli, Mousavian, Tyree, Birchfield, Tremblay, Carpentier, Aubry, Fox, and Sivic]{labbe2022megapose}
Yann Labb{\'{e}}, Lucas Manuelli, Arsalan Mousavian, Stephen Tyree, Stan Birchfield, Jonathan Tremblay, Justin Carpentier, Mathieu Aubry, Dieter Fox, and Josef Sivic.
\newblock {MegaPose}: {6D} pose estimation of novel objects via render {\&} compare.
\newblock In \emph{Conference on Robot Learning, CoRL 2022, 14-18 December 2022, Auckland, New Zealand}, pages 715--725. {PMLR}, 2023.

\bibitem[Li et~al.(2025)Li, Li, Liu, Li, and Huang]{li2025maniptrans}
Kailin Li, Puhao Li, Tengyu Liu, Yuyang Li, and Siyuan Huang.
\newblock Maniptrans: Efficient dexterous bimanual manipulation transfer via residual learning.
\newblock In \emph{Proceedings of the IEEE/CVF Conference on Computer Vision and Pattern Recognition (CVPR)}, pages 6991--7003, 2025.

\bibitem[Li et~al.(2026{\natexlab{a}})Li, Li, Deng, Xu, Yang, and Guo]{li2026vjepahand}
Ruicheng Li, Qixiu Li, Yu Deng, Sicheng Xu, Jiaolong Yang, and Baining Guo.
\newblock {V-JEPA Hand}: Simple, accurate and fast {3D} hand and camera trajectory recovery from egocentric videos.
\newblock \emph{arXiv preprint arXiv:2610.26780}, 2026{\natexlab{a}}.

\bibitem[Li et~al.(2026{\natexlab{b}})Li, Cho, Li, Guo, Wen, and Zhang]{li2026hof}
Shiqi Li, Sean Cho, Yijie Li, Fengzhi Guo, Bowen Wen, and Cheng Zhang.
\newblock {4D-HOF}: Hand-object flow matching for feed-forward {4D} interaction reconstruction.
\newblock \emph{arXiv preprint arXiv:2610.08782}, 2026{\natexlab{b}}.

\bibitem[Lin et~al.(2026)Lin, Chen, Liew, Chen, Li, Shi, Feng, and Kang]{lin2025depthanything3}
Haotong Lin, Sili Chen, Junhao Liew, Donny~Y. Chen, Zhenyu Li, Guang Shi, Jiashi Feng, and Bingyi Kang.
\newblock {Depth Anything 3}: Recovering the visual space from any views.
\newblock In \emph{International Conference on Learning Representations (ICLR)}, 2026.

\bibitem[Liu and Yi(2024)]{liu2024geneoh}
Xueyi Liu and Li Yi.
\newblock {GeneOH} diffusion: Towards generalizable hand-object interaction denoising via denoising diffusion.
\newblock In \emph{International Conference on Learning Representations (ICLR)}, 2024.

\bibitem[Liu et~al.(2024)Liu, Lyu, Zhang, Du, and Yi]{liu2024quasisim}
Xueyi Liu, Kangbo Lyu, Jieqiong Zhang, Tao Du, and Li Yi.
\newblock {QuasiSim}: Parameterized quasi-physical simulators for dexterous manipulations transfer.
\newblock In \emph{Computer Vision -- ECCV 2024}, pages 164--182. Springer Nature Switzerland, 2024.

\bibitem[Liu et~al.(2025{\natexlab{a}})Liu, Adalibieke, Han, Qin, and Yi]{liu2025dextrack}
Xueyi Liu, Jianibieke Adalibieke, Qianwei Han, Yuzhe Qin, and Li Yi.
\newblock {DexTrack}: Towards generalizable neural tracking control for dexterous manipulation from human references.
\newblock In \emph{International Conference on Learning Representations}, pages 85970--85994, 2025{\natexlab{a}}.

\bibitem[Liu et~al.(2025{\natexlab{b}})Liu, Long, Yang, Liu, Habermann, Theobalt, Ma, and Wang]{liu2025easyhoi}
Yumeng Liu, Xiaoxiao Long, Zemin Yang, Yuan Liu, Marc Habermann, Christian Theobalt, Yuexin Ma, and Wenping Wang.
\newblock {EasyHOI}: Unleashing the power of large models for reconstructing hand-object interactions in the wild.
\newblock In \emph{Proceedings of the IEEE/CVF Conference on Computer Vision and Pattern Recognition (CVPR)}, pages 7037--7047, 2025{\natexlab{b}}.

\bibitem[Luo et~al.(2024)Luo, Liu, and Yi]{luo2024physicshoi}
Haowen Luo, Yunze Liu, and Li Yi.
\newblock Physics-aware hand-object interaction denoising.
\newblock In \emph{{IEEE/CVF} Conference on Computer Vision and Pattern Recognition, {CVPR} 2024, Seattle, WA, USA, June 16-22, 2024}, pages 2341--2350. {IEEE}, 2024.

\bibitem[Mao et~al.(2026)Mao, Shao, Wang, Liu, Wang, Wu, and Mu]{mao2026omnihoi}
Ting Mao, Yanming Shao, Ziheng Wang, Haoyu Liu, Yiqun Wang, Xuanye Wu, and Yao Mu.
\newblock {OmniHOI}: Dexterous hand-object interaction from monocular human video.
\newblock \emph{arXiv preprint arXiv:2610.10855}, 2026.

\bibitem[On et~al.(2025)On, Gwak, Kang, Cha, Hwang, Hwang, and Baek]{on2025bigs}
Jeongwan On, Kyeonghwan Gwak, Gunyoung Kang, Junuk Cha, Soohyun Hwang, Hyein Hwang, and Seungryul Baek.
\newblock {BIGS}: Bimanual category-agnostic interaction reconstruction from monocular videos via {3D} gaussian splatting.
\newblock In \emph{Proceedings of the IEEE/CVF Conference on Computer Vision and Pattern Recognition (CVPR)}, pages 17437--17447, 2025.

\bibitem[Oquab et~al.(2024)Oquab, Darcet, Moutakanni, Vo, Szafraniec, Khalidov, Fernandez, Haziza, Massa, El-Nouby, Assran, Ballas, Galuba, Howes, Huang, Li, Misra, Rabbat, Sharma, Synnaeve, Xu, Jegou, Mairal, Labatut, Joulin, and Bojanowski]{oquab2024dinov2}
Maxime Oquab, Timoth{\'e}e Darcet, Th{\'e}o Moutakanni, Huy~V. Vo, Marc Szafraniec, Vasil Khalidov, Pierre Fernandez, Daniel Haziza, Francisco Massa, Alaaeldin El-Nouby, Mido Assran, Nicolas Ballas, Wojciech Galuba, Russell Howes, Po-Yao Huang, Shang-Wen Li, Ishan Misra, Michael Rabbat, Vasu Sharma, Gabriel Synnaeve, Hu Xu, Herve Jegou, Julien Mairal, Patrick Labatut, Armand Joulin, and Piotr Bojanowski.
\newblock {DINOv2}: Learning robust visual features without supervision.
\newblock \emph{Transactions on Machine Learning Research}, 2024.

\bibitem[{\"{O}}rnek et~al.(2024){\"{O}}rnek, Labb{\'{e}}, Tekin, Ma, Keskin, Forster, and Hodan]{ornek2024foundpose}
Evin~Pinar {\"{O}}rnek, Yann Labb{\'{e}}, Bugra Tekin, Lingni Ma, Cem Keskin, Christian Forster, and Tomas Hodan.
\newblock {FoundPose}: Unseen object pose estimation with foundation features.
\newblock In \emph{Computer Vision - {ECCV} 2024 - 18th European Conference, Milan, Italy, September 29-October 4, 2024, Proceedings, Part {XXVI}}, pages 163--182. Springer, 2024.

\bibitem[Paliwal et~al.(2026)Paliwal, Etukuru, Liang, Abbeel, Shafiullah, and Malik]{paliwal2026doasido}
Bhawna Paliwal, Haritheja Etukuru, William Liang, Pieter Abbeel, Nur Muhammad~Mahi Shafiullah, and Jitendra Malik.
\newblock Do as i do: Dexterous manipulation data from everyday human videos.
\newblock \emph{arXiv preprint arXiv:2606.19333}, 2026.

\bibitem[Pan et~al.(2026)Pan, Wang, Qi, Urain, Liu, Bharadhwaj, Sharma, Wu, Shi, Malik, and Hogan]{pan2026spider}
Chaoyi Pan, Changhao Wang, Haozhi Qi, Julen Urain, Zixi Liu, Homanga Bharadhwaj, Akash Sharma, Tingfan Wu, Guanya Shi, Jitendra Malik, and Francois Hogan.
\newblock {SPIDER}: Scalable physics-informed dexterous retargeting.
\newblock \emph{arXiv preprint arXiv:2511.09484}, 2026.

\bibitem[Park et~al.(2025)Park, Lee, Choi, Lee, Kim, Kim, and Joo]{park2025learning}
Sungjae Park, Seungho Lee, Mingi Choi, Jiye Lee, Jeonghwan Kim, Jisoo Kim, and Hanbyul Joo.
\newblock Learning to transfer human hand skills for robot manipulations.
\newblock \emph{arXiv preprint arXiv:2501.04169}, 2025.

\bibitem[Pavlakos et~al.(2024)Pavlakos, Shan, Radosavovic, Kanazawa, Fouhey, and Malik]{pavlakos2024hamer}
Georgios Pavlakos, Dandan Shan, Ilija Radosavovic, Angjoo Kanazawa, David Fouhey, and Jitendra Malik.
\newblock Reconstructing hands in {3D} with transformers.
\newblock In \emph{{IEEE/CVF} Conference on Computer Vision and Pattern Recognition, {CVPR} 2024, Seattle, WA, USA, June 16-22, 2024}, pages 9826--9836. {IEEE}, 2024.

\bibitem[Peebles and Xie(2023)]{peebles2023dit}
William Peebles and Saining Xie.
\newblock Scalable diffusion models with transformers.
\newblock In \emph{Proceedings of the IEEE/CVF International Conference on Computer Vision (ICCV)}, pages 4195--4205, 2023.

\bibitem[Ponimatkin et~al.(2025)Ponimatkin, C\'{\i}fka, Soucek, Fourmy, Labb\'{e}, Petrik, and Sivic]{ponimatkin2025object}
Georgy Ponimatkin, Martin C\'{\i}fka, Tomas Soucek, M\'{e}d\'{e}ric Fourmy, Yann Labb\'{e}, Vladimir Petrik, and Josef Sivic.
\newblock {6D} object pose tracking in {Internet} videos for robotic manipulation.
\newblock In \emph{International Conference on Learning Representations}, pages 32569--32596, 2025.

\bibitem[Potamias et~al.(2025)Potamias, Zhang, Deng, and Zafeiriou]{potamias2025wilor}
Rolandos~Alexandros Potamias, Jinglei Zhang, Jiankang Deng, and Stefanos Zafeiriou.
\newblock {WiLoR}: End-to-end {3D} hand localization and reconstruction in-the-wild.
\newblock In \emph{{IEEE/CVF} Conference on Computer Vision and Pattern Recognition, {CVPR} 2025, Nashville, TN, USA, June 11-15, 2025}, pages 12242--12254. Computer Vision Foundation / {IEEE}, 2025.

\bibitem[Prokudin et~al.(2019)Prokudin, Lassner, and Romero]{prokudin2019bps}
Sergey Prokudin, Christoph Lassner, and Javier Romero.
\newblock Efficient learning on point clouds with basis point sets.
\newblock In \emph{Proceedings of the IEEE/CVF International Conference on Computer Vision (ICCV)}, 2019.

\bibitem[Qin et~al.(2022)Qin, Wu, Liu, Jiang, Yang, Fu, and Wang]{qin2022dexmv}
Yuzhe Qin, Yueh{-}Hua Wu, Shaowei Liu, Hanwen Jiang, Ruihan Yang, Yang Fu, and Xiaolong Wang.
\newblock {DexMV}: Imitation learning for dexterous manipulation from human videos.
\newblock In \emph{Computer Vision - {ECCV} 2022 - 17th European Conference, Tel Aviv, Israel, October 23-27, 2022, Proceedings, Part {XXXIX}}, pages 570--587. Springer, 2022.

\bibitem[Romero et~al.(2017)Romero, Tzionas, and Black]{romero2017mano}
Javier Romero, Dimitrios Tzionas, and Michael~J. Black.
\newblock Embodied hands: modeling and capturing hands and bodies together.
\newblock \emph{{ACM} Trans. Graph.}, 36\penalty0 (6):\penalty0 245:1--245:17, 2017.

\bibitem[Shaw et~al.(2023)Shaw, Bahl, and Pathak]{shaw2022videodex}
Kenneth Shaw, Shikhar Bahl, and Deepak Pathak.
\newblock {VideoDex}: Learning dexterity from {Internet} videos.
\newblock In \emph{Conference on Robot Learning, CoRL 2022, 14-18 December 2022, Auckland, New Zealand}, pages 654--665. {PMLR}, 2023.

\bibitem[Shin et~al.(2026)Shin, Lee, Lee, Bae, Lee, Im, Lee, and Jeon]{shin2026compose}
Jisu Shin, Junoh Lee, JunGyu Lee, Inhwan Bae, Dohyeon Lee, Hokyun Im, Youngwoon Lee, and Hae-Gon Jeon.
\newblock {ComPose}: When to trust hands for object pose tracking.
\newblock \emph{arXiv preprint arXiv:2605.23523}, 2026.

\bibitem[Wang et~al.(2026{\natexlab{a}})Wang, Ren, Morgan, Chen, Qian, Chanrungmaneekul, and Hang]{wang2026egoinfinity}
Gaotian Wang, Kejia Ren, Andrew Morgan, Yiting Chen, Howard~H. Qian, Podshara Chanrungmaneekul, and Kaiyu Hang.
\newblock {EgoInfinity}: A web-scale {4D} hand-object interaction data engine for any-view robot retargeting and video-to-action robot learning.
\newblock \emph{arXiv preprint arXiv:2606.17385}, 2026{\natexlab{a}}.

\bibitem[Wang et~al.(2025)Wang, He, Parelli, Gebhardt, Fan, and Song]{wang2025magichoi}
Shibo Wang, Haonan He, Maria Parelli, Christoph Gebhardt, Zicong Fan, and Jie Song.
\newblock {MagicHOI}: Leveraging {3D} priors for accurate hand-object reconstruction from short monocular video clips.
\newblock In \emph{Proceedings of the IEEE/CVF International Conference on Computer Vision (ICCV)}, pages 5957--5968, 2025.

\bibitem[Wang et~al.(2026{\natexlab{b}})Wang, Zhou, Zhu, Chang, Zhou, Li, Chen, Pang, Shen, and He]{wang2025pi3}
Yifan Wang, Jianjun Zhou, Haoyi Zhu, Wenzheng Chang, Yang Zhou, Zizun Li, Junyi Chen, Jiangmiao Pang, Chunhua Shen, and Tong He.
\newblock {$\pi^3$}: Permutation-equivariant visual geometry learning.
\newblock In \emph{International Conference on Learning Representations (ICLR)}, 2026{\natexlab{b}}.

\bibitem[Wen and Bekris(2021)]{wen2021bundletrack}
Bowen Wen and Kostas Bekris.
\newblock {BundleTrack}: {6D} pose tracking for novel objects without instance or category-level {3D} models.
\newblock In \emph{2021 IEEE/RSJ International Conference on Intelligent Robots and Systems (IROS)}, pages 8067--8074. IEEE, 2021.

\bibitem[Wen et~al.(2023)Wen, Tremblay, Blukis, Tyree, M\"uller, Evans, Fox, Kautz, and Birchfield]{wen2023bundlesdf}
Bowen Wen, Jonathan Tremblay, Valts Blukis, Stephen Tyree, Thomas M\"uller, Alex Evans, Dieter Fox, Jan Kautz, and Stan Birchfield.
\newblock {BundleSDF}: Neural {6-DoF} tracking and {3D} reconstruction of unknown objects.
\newblock In \emph{Proceedings of the IEEE/CVF Conference on Computer Vision and Pattern Recognition (CVPR)}, pages 606--617, 2023.

\bibitem[Wen et~al.(2024)Wen, Yang, Kautz, and Birchfield]{wen2024foundationpose}
Bowen Wen, Wei Yang, Jan Kautz, and Stan Birchfield.
\newblock {FoundationPose}: Unified {6D} pose estimation and tracking of novel objects.
\newblock In \emph{{IEEE/CVF} Conference on Computer Vision and Pattern Recognition, {CVPR} 2024, Seattle, WA, USA, June 16-22, 2024}, pages 17868--17879. {IEEE}, 2024.

\bibitem[Xu et~al.(2026)Xu, Liu, Wang, Fu, and Mitra]{xu2026openchoir}
Hao Xu, Yilin Liu, Yinqiao Wang, Chi-Wing Fu, and Niloy~J. Mitra.
\newblock {Open-CHOIR}: Open-world contact-aware {4D} hand-object interaction reconstruction.
\newblock \emph{ACM Transactions on Graphics}, pages 1--12, 2026.

\bibitem[Ye et~al.(2022)Ye, Gupta, and Tulsiani]{ye2022ihoi}
Yufei Ye, Abhinav Gupta, and Shubham Tulsiani.
\newblock What's in your hands? {3D} reconstruction of generic objects in hands.
\newblock In \emph{{IEEE/CVF} Conference on Computer Vision and Pattern Recognition, {CVPR} 2022, New Orleans, LA, USA, June 18-24, 2022}, pages 3885--3895. {IEEE}, 2022.

\bibitem[Ye et~al.(2023)Ye, Hebbar, Gupta, and Tulsiani]{ye2023diffhoi}
Yufei Ye, Poorvi Hebbar, Abhinav Gupta, and Shubham Tulsiani.
\newblock Diffusion-guided reconstruction of everyday hand-object interaction clips.
\newblock In \emph{{IEEE/CVF} International Conference on Computer Vision, {ICCV} 2023, Paris, France, October 1-6, 2023}, pages 19660--19671. {IEEE}, 2023.

\bibitem[Ye et~al.(2024)Ye, Gupta, Kitani, and Tulsiani]{ye2024ghop}
Yufei Ye, Abhinav Gupta, Kris Kitani, and Shubham Tulsiani.
\newblock {G-HOP:} generative hand-object prior for interaction reconstruction and grasp synthesis.
\newblock In \emph{{IEEE/CVF} Conference on Computer Vision and Pattern Recognition, {CVPR} 2024, Seattle, WA, USA, June 16-22, 2024}, pages 1911--1920. {IEEE}, 2024.

\bibitem[Ye et~al.(2026{\natexlab{a}})Ye, Feng, Taheri, Feng, Tulsiani, and Black]{ye2026haptic}
Yufei Ye, Yao Feng, Omid Taheri, Haiwen Feng, Shubham Tulsiani, and Michael~J. Black.
\newblock Predicting {4D} hand trajectory from monocular videos.
\newblock In \emph{2026 International Conference on 3D Vision (3DV)}, pages 860--870. IEEE, 2026{\natexlab{a}}.

\bibitem[Ye et~al.(2026{\natexlab{b}})Ye, Li, Rong, and Liu]{ye2026whole}
Yufei Ye, Jiaman Li, Ryan Rong, and C.~Karen Liu.
\newblock {WHOLE}: World-grounded hand-object lifted from egocentric videos.
\newblock In \emph{Proceedings of the IEEE/CVF Conference on Computer Vision and Pattern Recognition (CVPR) Findings}, pages 3481--3491, 2026{\natexlab{b}}.

\bibitem[Yu et~al.(2025)Yu, Zafeiriou, and Birdal]{yu2025dynhamr}
Zhengdi Yu, Stefanos Zafeiriou, and Tolga Birdal.
\newblock {Dyn-HaMR}: Recovering {4D} interacting hand motion from a dynamic camera.
\newblock In \emph{{IEEE/CVF} Conference on Computer Vision and Pattern Recognition, {CVPR} 2025, Nashville, TN, USA, June 11-15, 2025}, pages 27716--27726. Computer Vision Foundation / {IEEE}, 2025.

\bibitem[Zeng et~al.(2026)Zeng, Yao, Zhang, Yang, Potamias, and Zuo]{zeng2026stablehand}
Huajian Zeng, Chaohua Yao, Yuantai Zhang, Jiaqi Yang, Rolandos~Alexandros Potamias, and Xingxing Zuo.
\newblock {StableHand}: Quality-aware flow matching for world-space dual-hand motion estimation from egocentric video.
\newblock \emph{arXiv preprint arXiv:2605.18553}, 2026.

\bibitem[Zhang et~al.(2025)Zhang, Deng, Ma, and Potamias]{zhang2025hawor}
Jinglei Zhang, Jiankang Deng, Chao Ma, and Rolandos~Alexandros Potamias.
\newblock {HaWoR}: World-space hand motion reconstruction from egocentric videos.
\newblock In \emph{{IEEE/CVF} Conference on Computer Vision and Pattern Recognition, {CVPR} 2025, Nashville, TN, USA, June 11-15, 2025}, pages 1805--1815. Computer Vision Foundation / {IEEE}, 2025.

\bibitem[Zhang et~al.(2026)Zhang, Gao, Li, Chen, Liu, Cheng, Lin, Liu, Li, Feng, He, Lin, Huang, Liu, and Wang]{zhang2025robowheel}
Yuhong Zhang, Zihan Gao, Shengpeng Li, Ling-Hao Chen, Kaisheng Liu, Runqing Cheng, Xiao Lin, Junjia Liu, Zhuoheng Li, Jingyi Feng, Ziyan He, Jintian Lin, Zheyan Huang, Zhifang Liu, and Haoqian Wang.
\newblock {RoboWheel}: A data engine from real-world human demonstrations for cross-embodiment robotic learning.
\newblock In \emph{Proceedings of the IEEE/CVF Conference on Computer Vision and Pattern Recognition (CVPR)}, pages 6664--6674, 2026.

\bibitem[Zhou et~al.(2019)Zhou, Barnes, Lu, Yang, and Li]{zhou2019continuity}
Yi Zhou, Connelly Barnes, Jingwan Lu, Jimei Yang, and Hao Li.
\newblock On the continuity of rotation representations in neural networks.
\newblock In \emph{Proceedings of the IEEE/CVF Conference on Computer Vision and Pattern Recognition (CVPR)}, 2019.

\bibitem[Zhu et~al.(2026{\natexlab{a}})Zhu, Liu, Jain, Li, Noori, Lin, Zhao, Welsh, Verghese, Liu, Wang, Da, Luo, Kulkarni, Bhatti, Zhu, Fan, Wen, Xu, Pouya, and Chang]{zhu2026chord}
Xinghao Zhu, Zixi Liu, Shalin Jain, Chenran Li, Milad Noori, Michael~Andres Lin, Huihua Zhao, John Welsh, Mrinal Verghese, Wei Liu, Tingwu Wang, Xingye Da, Zhengyi Luo, Vishal Kulkarni, Naema Bhatti, Yuke Zhu, Linxi Fan, Bowen Wen, Danfei Xu, Soha Pouya, and Yan Chang.
\newblock Learning dexterous manipulation using contact wrench guidance from human demonstration.
\newblock \emph{arXiv preprint arXiv:2607.00033}, 2026{\natexlab{a}}.

\bibitem[Zhu et~al.(2026{\natexlab{b}})Zhu, Bansal, Tripathi, and Damen]{zhu2026rohit}
Zhifan Zhu, Siddhant Bansal, Shashank Tripathi, and Dima Damen.
\newblock Reconstructing objects along hand interaction timelines in egocentric video.
\newblock In \emph{Proceedings of the IEEE/CVF Conference on Computer Vision and Pattern Recognition (CVPR) Workshops}, pages 3955--3965, 2026{\natexlab{b}}.

\end{thebibliography}
